\documentclass[10pt,twocolumn,letterpaper]{article}

\usepackage{cvpr}

\usepackage{times}

\usepackage{multirow}
\usepackage{rotating}
\usepackage{booktabs}
\usepackage{slashbox} 
\usepackage{algpseudocode}
\usepackage{algorithm}

\usepackage{epsfig}
\usepackage{graphicx}
\usepackage{amsmath}
\usepackage{amssymb}
\usepackage{caption}
\usepackage{soul,xcolor}
\usepackage{paralist}
\usepackage{soul}
\usepackage[normalem]{ulem}
\usepackage{xcolor}
\usepackage{subfig}

\usepackage{mysymbols}

\graphicspath{ {figures/} }

\usepackage[pagebackref=true,breaklinks=true,letterpaper=true,colorlinks,bookmarks=false]{hyperref}

\cvprfinalcopy 

\def\cvprPaperID{3177} 

\begin{document}

\title{Making the V in VQA Matter: \\
Elevating the Role of Image Understanding in Visual Question Answering}


\author{Yash Goyal\thanks{The first two authors contributed equally.} $\,^1$  
  \quad Tejas Khot\footnotemark[1] $\,^1$ \quad Douglas Summers-Stay$^2$ \quad Dhruv Batra$^3$ \quad Devi Parikh$^3$\\
$^1$Virginia Tech \quad $^2$Army Research Laboratory \quad $^3$Georgia Institute of Technology\\
{\tt\small $^1$\{ygoyal, tjskhot\}@vt.edu} \quad \tt\small $^2$douglas.a.summers-stay.civ@mail.mil \quad \tt\small $^3$\{dbatra, parikh\}@gatech.edu}

\maketitle

\begin{abstract}

Problems at the intersection of vision and language are of significant 
importance both as challenging 
research questions and for the rich set of applications they enable.  
However, inherent structure in our world and bias in our language 
tend to be a simpler signal for learning than visual modalities, 
resulting in models that ignore visual information, 
leading to an inflated sense of their capability. 

We propose to counter these language priors for the task of 
Visual Question Answering (VQA) and make vision (the V in VQA) matter! 
Specifically, we \emph{balance} the popular VQA dataset~\cite{VQA} by collecting 
complementary images such that every question 
in our balanced dataset is associated with not just a single image, but 
rather a \emph{pair of similar images} 
that result in two different answers to the question. 
Our dataset is by construction more balanced than the original VQA 
dataset and has approximately \emph{twice} 
the number of image-question pairs. 
Our complete balanced dataset is publicly available at \url{http://visualqa.org/} as part of the 2nd iteration of the Visual Question Answering Dataset and Challenge (VQA v2.0).

We further benchmark a number of state-of-art VQA models on our balanced dataset. 
All models perform significantly worse on our balanced dataset, 
suggesting that these models have indeed learned to exploit 
language priors. This finding provides the first concrete empirical evidence 
for what seems to be a qualitative sense among practitioners. 

Finally, our data collection protocol for identifying 
complementary images enables us to develop a novel interpretable model, 
which in addition to providing an answer to the given (image, question) 
pair, also provides a counter-example based explanation.
Specifically, it identifies an image that is similar to the original 
image, but it believes has a different answer to the same question. 
This can help in building trust for machines among their users.

\end{abstract}


\section{Introduction}
\label{sec:intro}

Language and vision problems such as image captioning 
\cite{captioning_msr, captioning_xinlei, captioning_berkeley, 
captioning_stanford, captioning_google, captioning_toronto, 
captioning_baidu_ucla} and visual question answering (VQA) 
\cite{VQA, fritz, Malinowski_2015_ICCV, baiduVQA, Ren_2015_NIPS} 
have gained popularity in recent years as the computer vision 
research community is progressing beyond ``bucketed'' recognition 
and towards solving multi-modal problems.

\begin{figure}[t]
\centering
\includegraphics[width=1\linewidth]{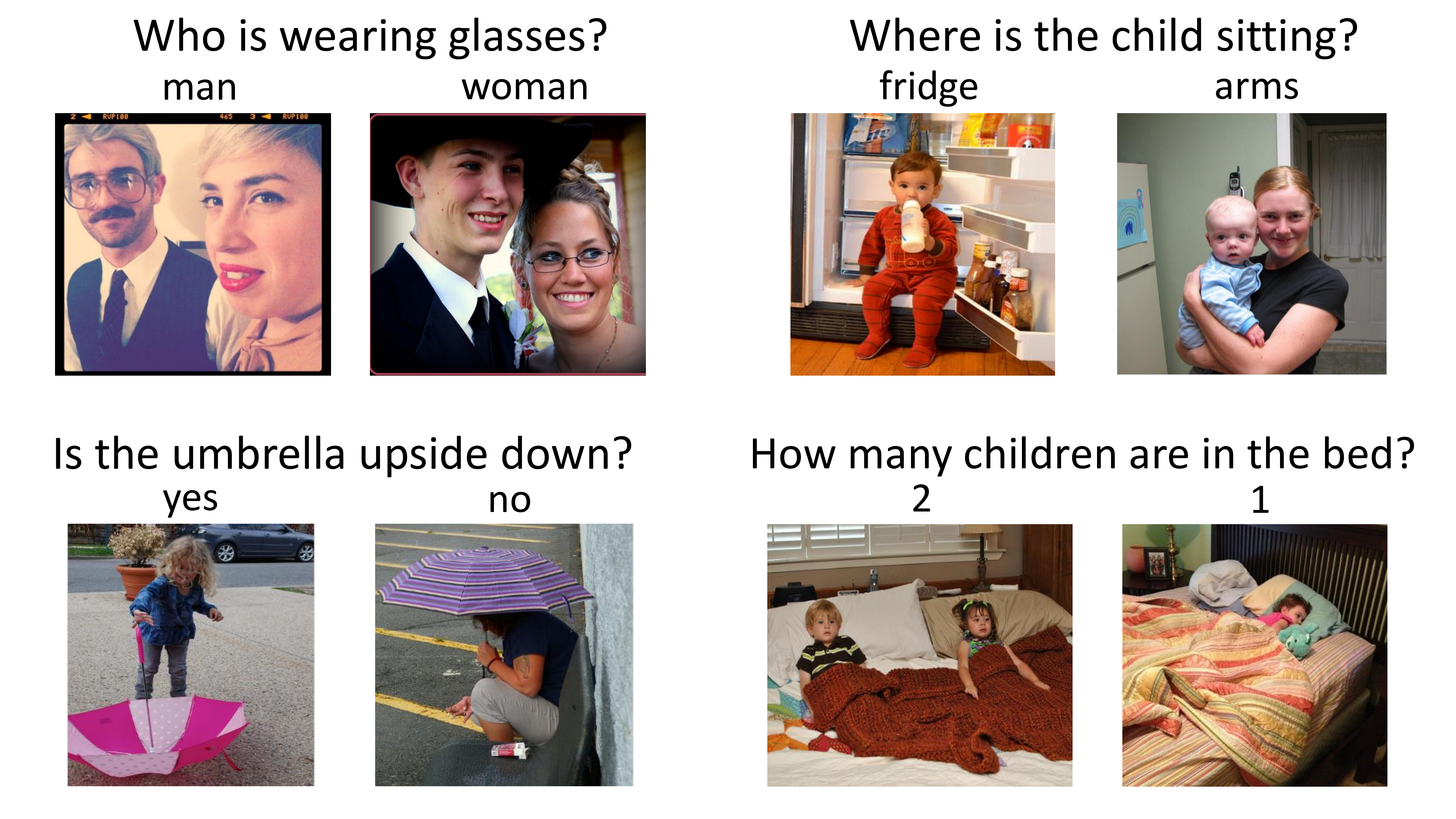}
\caption{Examples from our balanced VQA dataset.}
\label{fig:dataset}
\end{figure} 

The complex compositional structure of language makes problems 
at the intersection of vision and language challenging. 
But recent works \cite{Larry_NN_caption,binaryVQA,ZhouVQABaseline, 
FbVQA,Kushal_VQA,Aish_analyze} 
have pointed out that language also provides a strong prior 
that can result in good superficial performance, 
without the underlying models truly understanding the visual content. 

This phenomenon has been observed in image captioning \cite{Larry_NN_caption} 
as well as visual question answering \cite{binaryVQA,ZhouVQABaseline,FbVQA,Kushal_VQA,Aish_analyze}.
For instance, in the VQA \cite{VQA} dataset, the most common sport answer 
``tennis'' is the correct answer for 41\% of the questions starting with 
``What sport is'', and ``2'' is the correct answer for 39\% of the questions 
starting with ``How many''. 
Moreover, Zhang~\etal~\cite{binaryVQA} points out a particular `visual priming bias' 
in the VQA dataset -- 
specifically, subjects saw an image while asking questions about it. 
Thus, people only ask the question ``Is there a clock tower in the picture?'' on images 
actually containing clock towers. 
As one particularly perverse example -- for questions in the VQA dataset 
starting with the n-gram ``Do you see a \ldots'', 
blindly answering ``yes'' without reading the rest of the question 
or looking at the associated image results in a VQA accuracy of $87\%$!

These language priors can give a false impression that machines are making 
progress towards the goal of understanding images correctly when they are 
only exploiting language priors to achieve high accuracy. 
This can hinder progress in pushing state of art in the computer vision 
aspects of multi-modal AI \cite{DatasetBias,binaryVQA}. 

\begin{figure*}
\begin{tabular}{cccc}
\includegraphics[width = 0.22\linewidth]{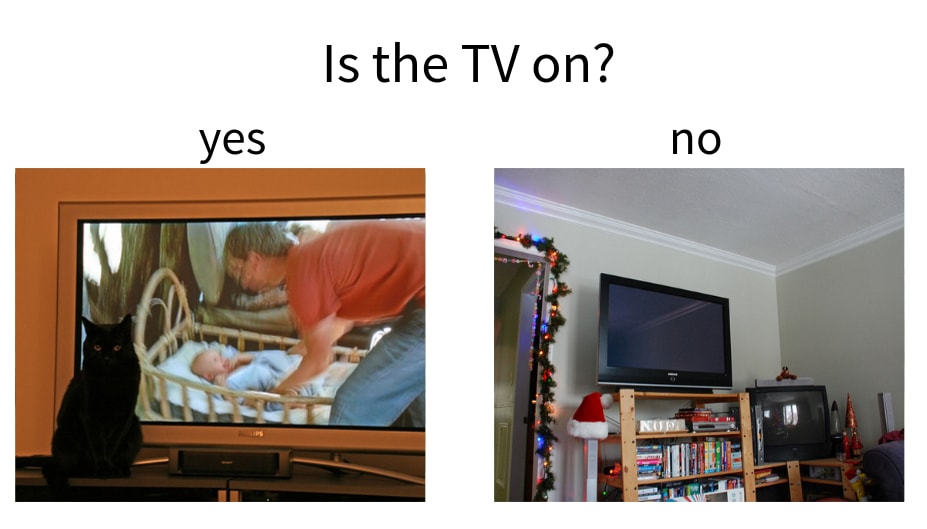} &
\includegraphics[width = 0.22\linewidth]{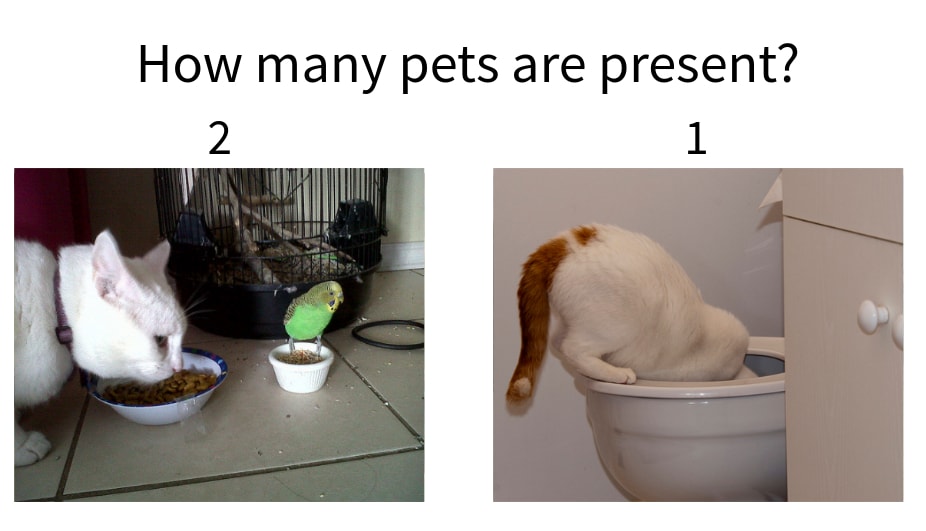} &
\includegraphics[width = 0.22\linewidth]{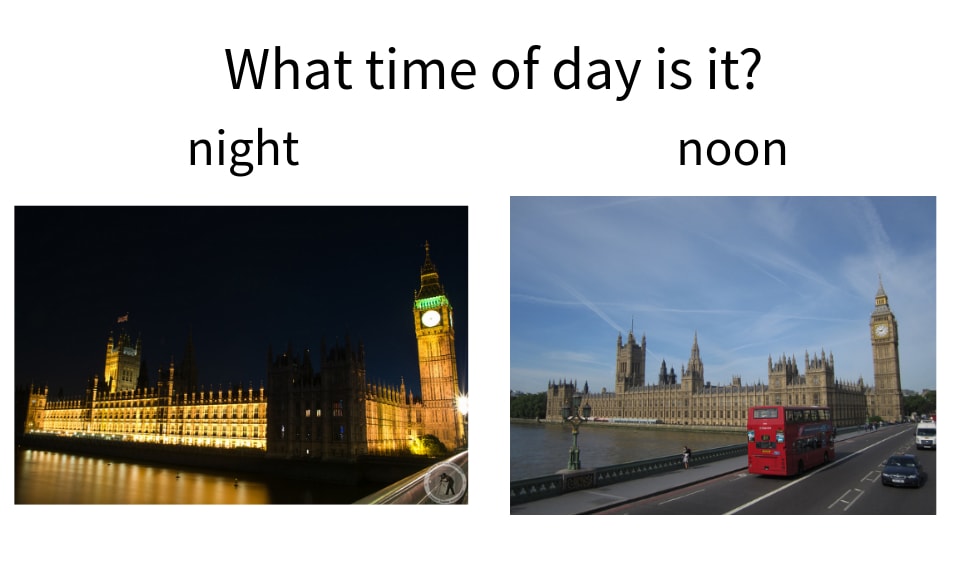} &
\includegraphics[width = 0.22\linewidth]{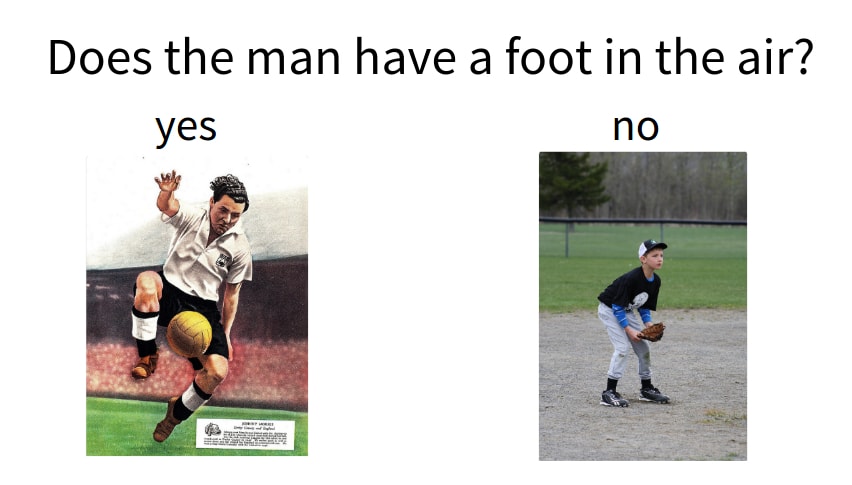}\\
\includegraphics[width = 0.22\linewidth]{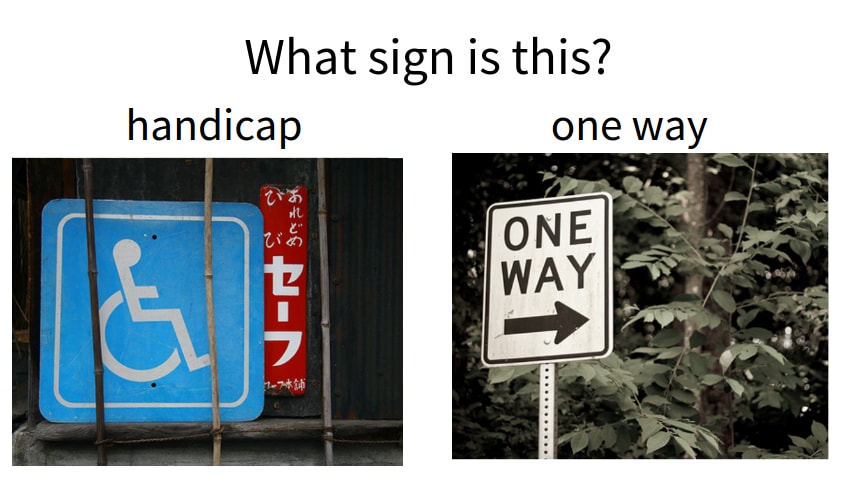} &
\includegraphics[width = 0.22\linewidth]{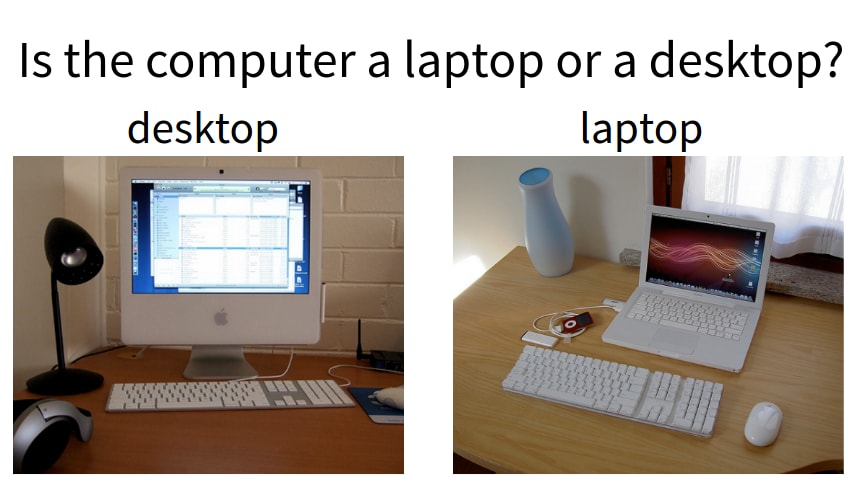} &
\includegraphics[width = 0.22\linewidth]{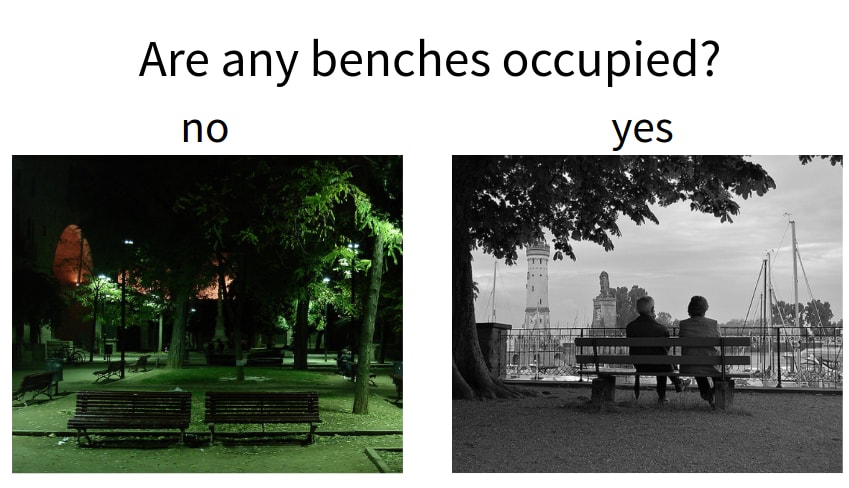} &
\includegraphics[width = 0.22\linewidth]{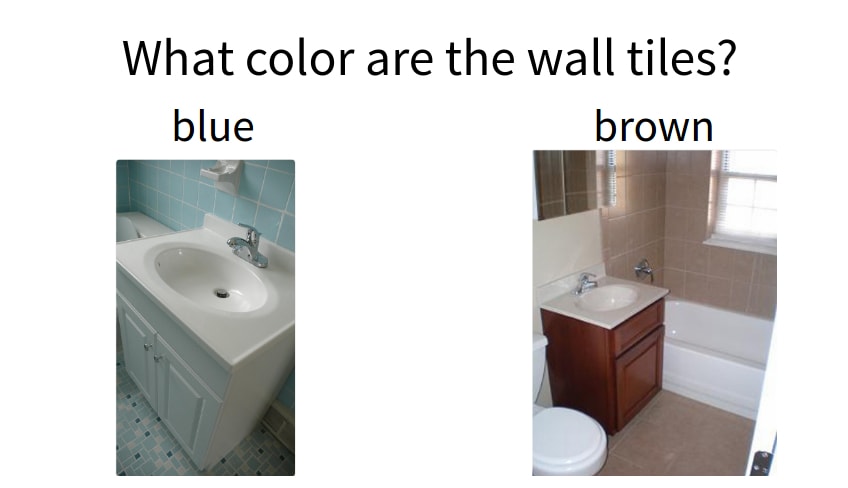}\\
\includegraphics[width = 0.22\linewidth]{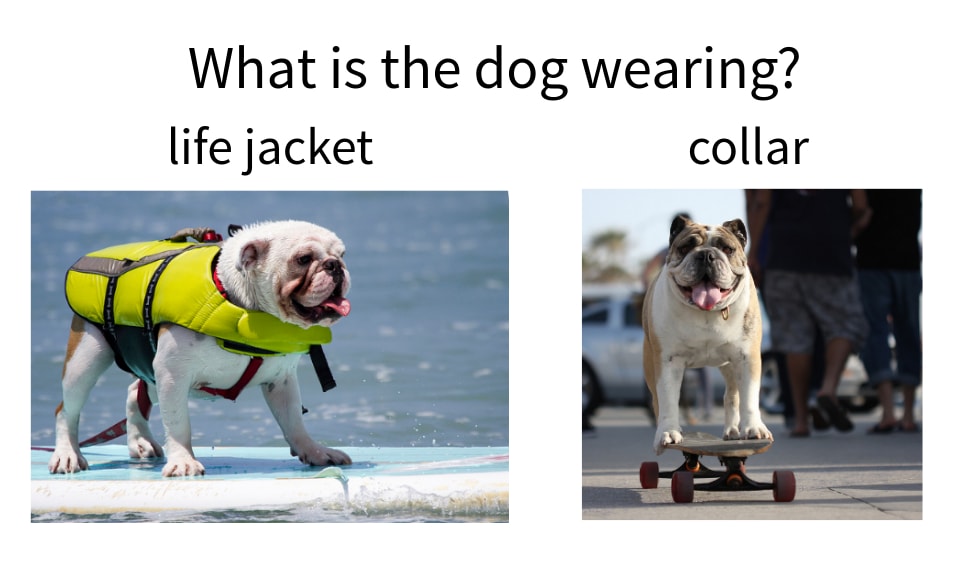} &
\includegraphics[width = 0.22\linewidth]{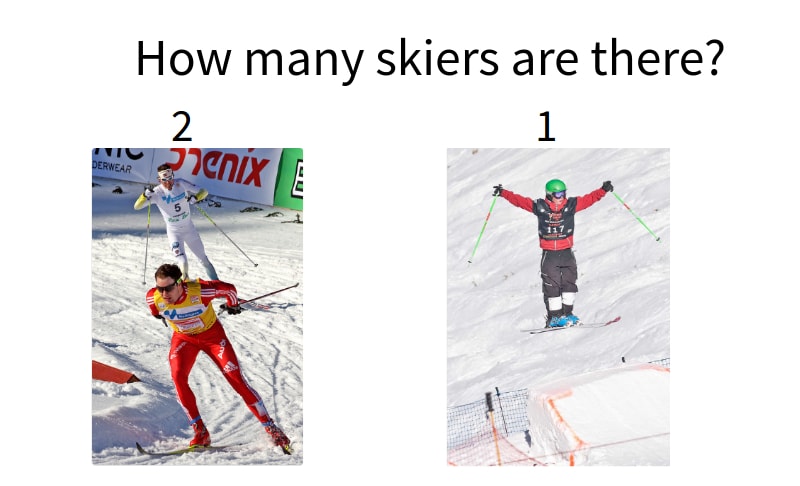} &
\includegraphics[width = 0.22\linewidth]{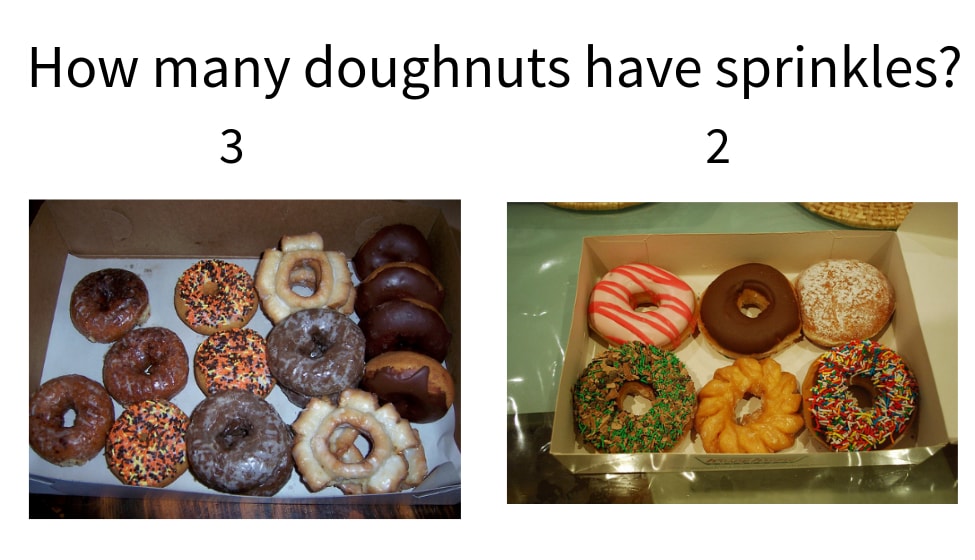} &
\includegraphics[width = 0.22\linewidth]{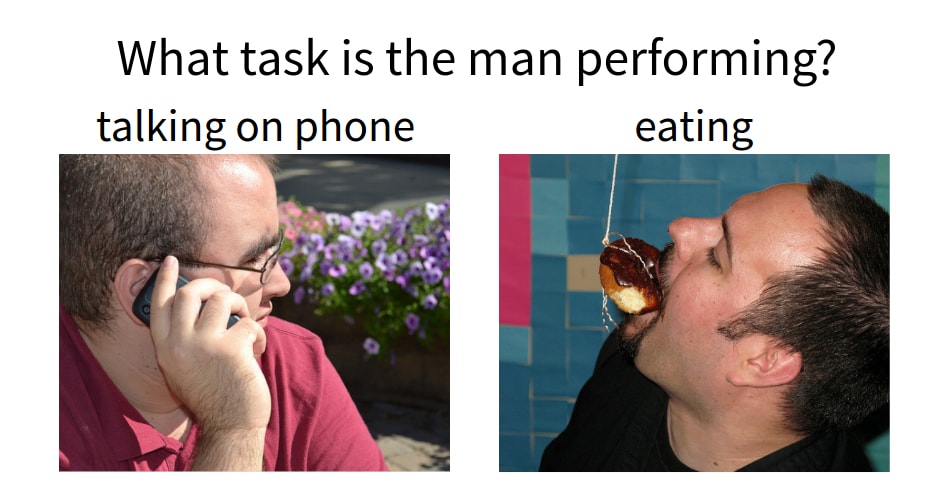}\\
\includegraphics[width = 0.22\linewidth]{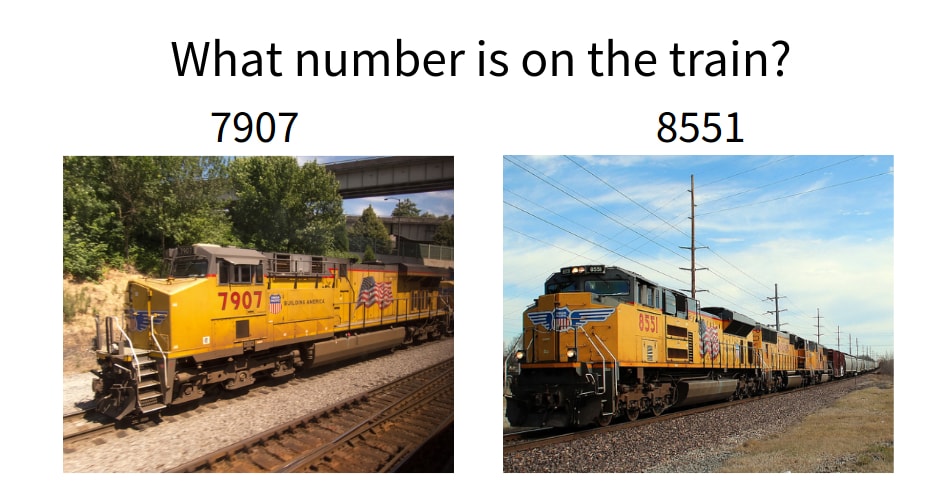} &
\includegraphics[width = 0.22\linewidth]{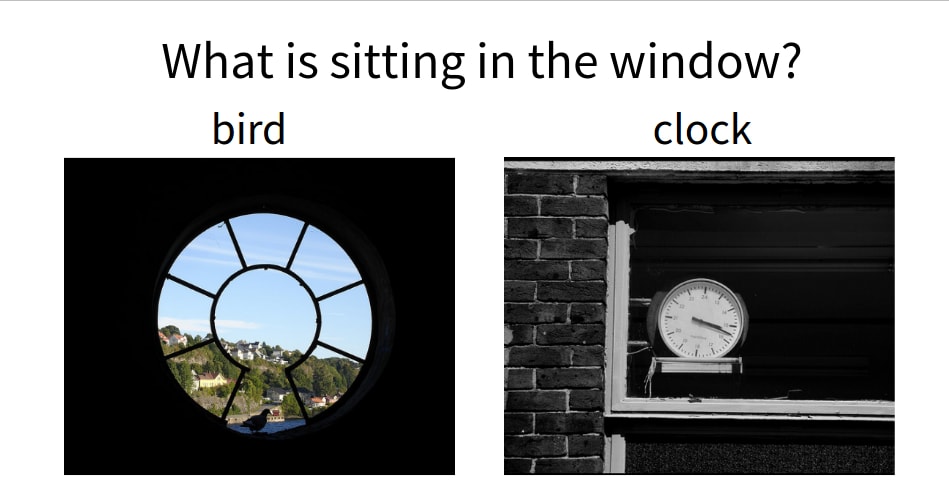} &
\includegraphics[width = 0.22\linewidth]{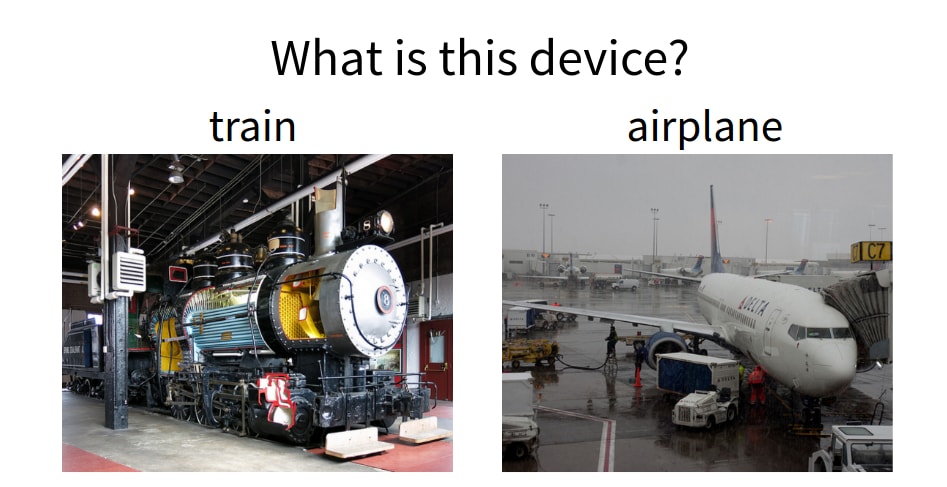} &
\includegraphics[width = 0.22\linewidth]{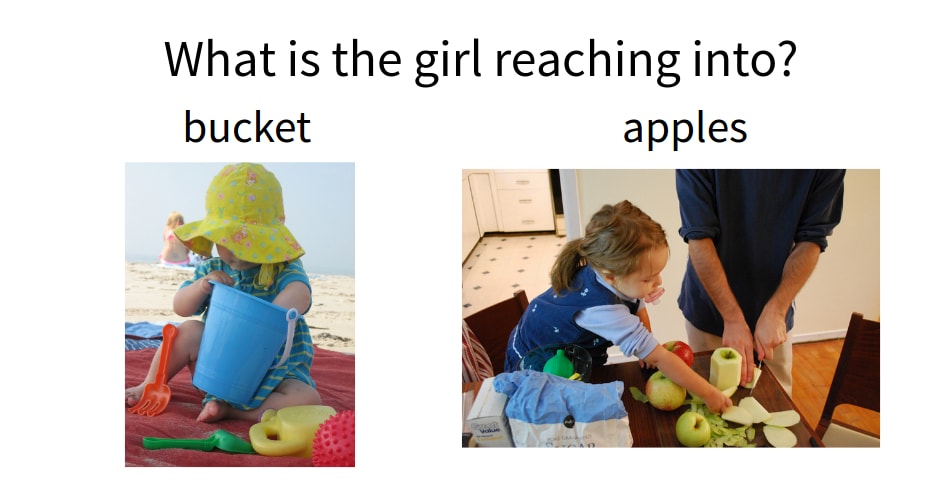}\\
\includegraphics[width = 0.22\linewidth]{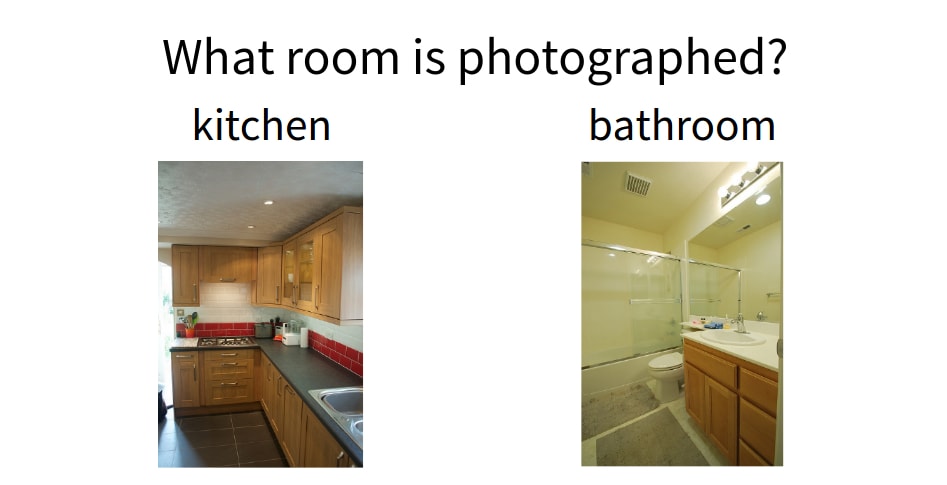} &
\includegraphics[width = 0.22\linewidth]{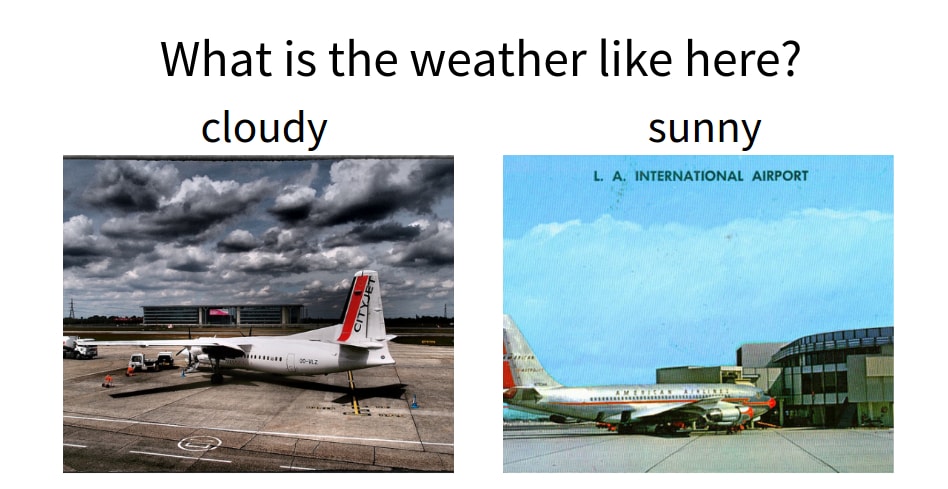} &
\includegraphics[width = 0.22\linewidth]{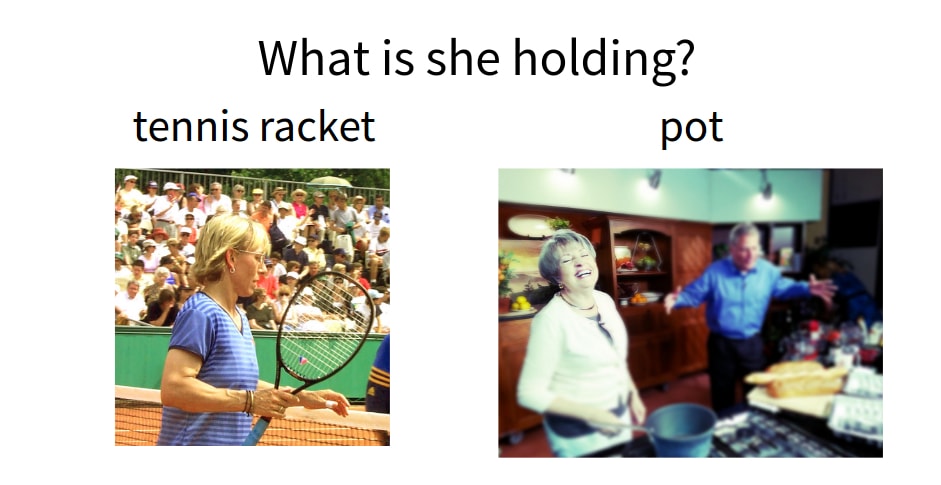} &
\includegraphics[width = 0.22\linewidth]{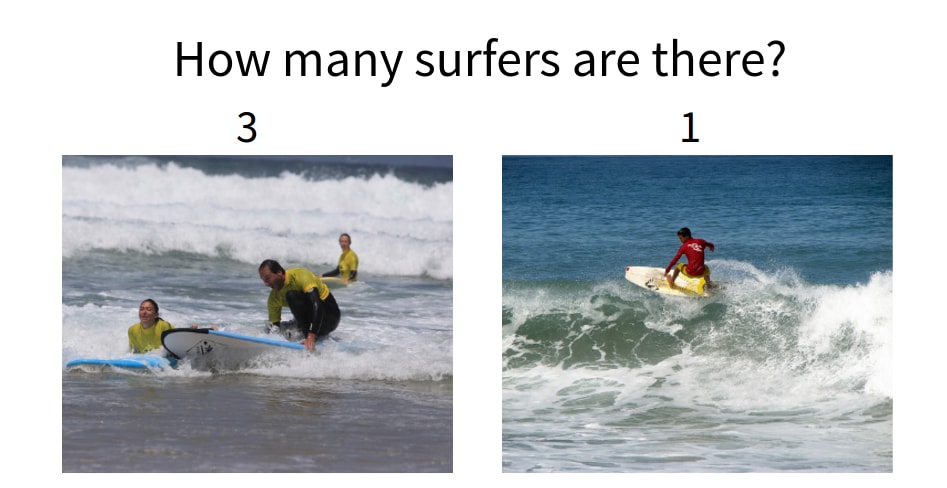}\\
\includegraphics[width = 0.22\linewidth]{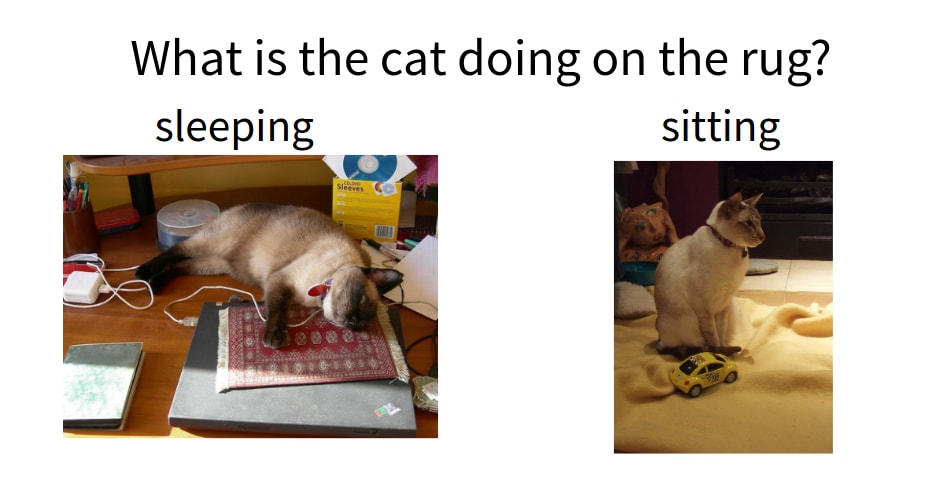} &
\includegraphics[width = 0.22\linewidth]{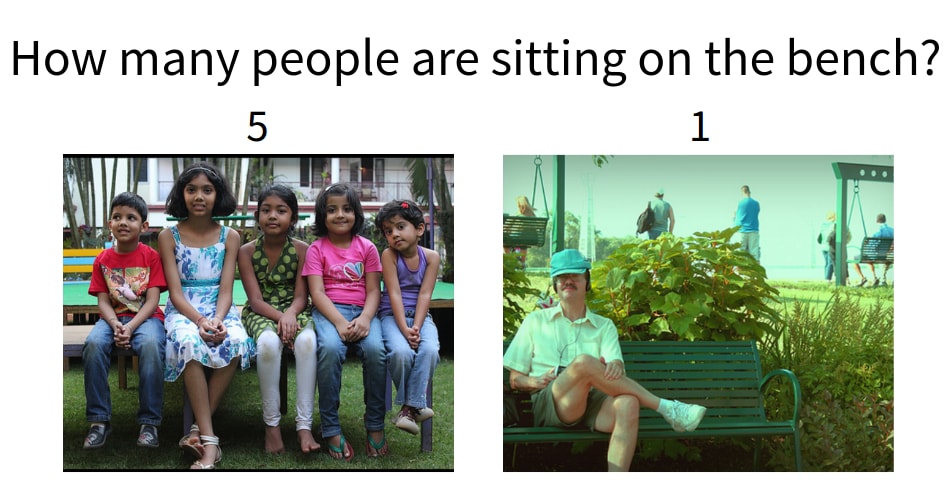} &
\includegraphics[width = 0.22\linewidth]{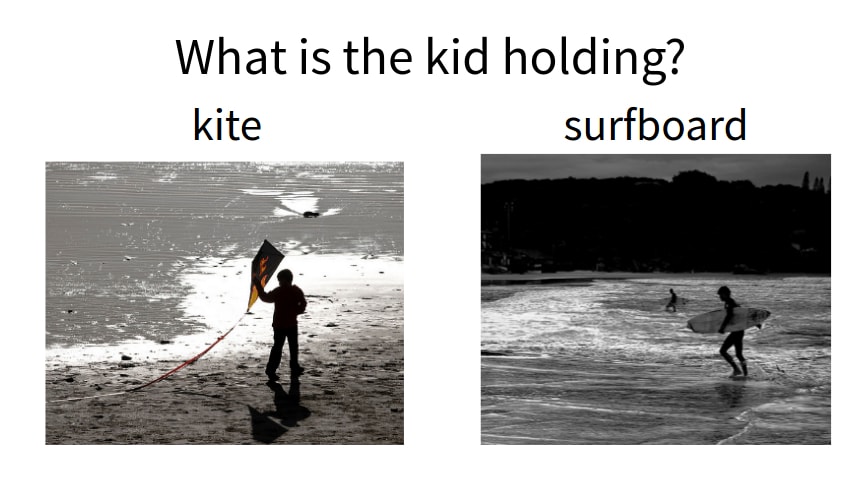} &
\includegraphics[width = 0.22\linewidth]{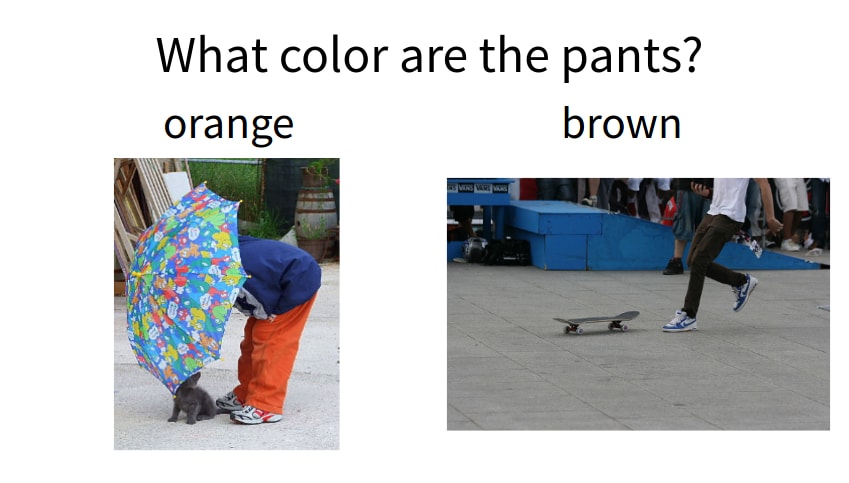}\\
\includegraphics[width = 0.22\linewidth]{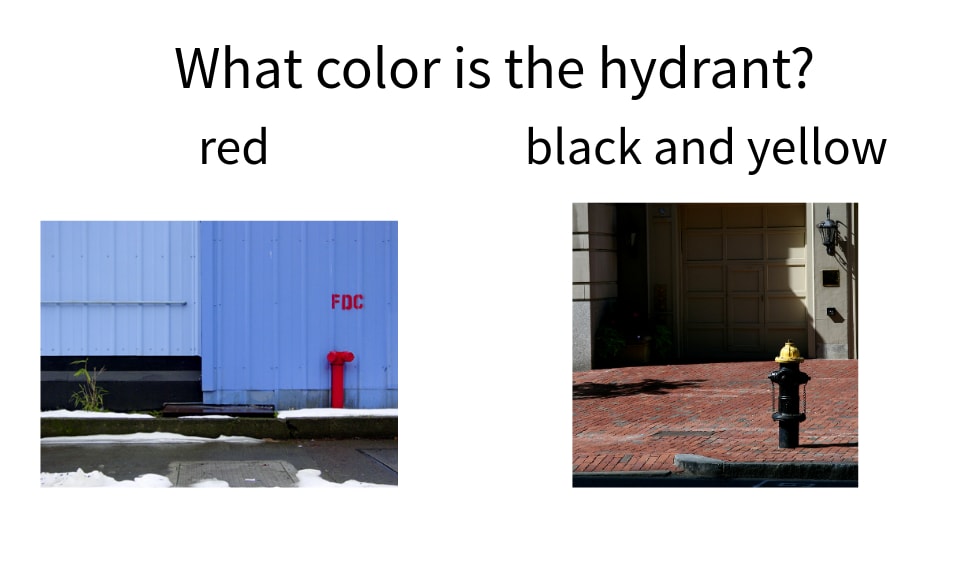} &
\includegraphics[width = 0.22\linewidth]{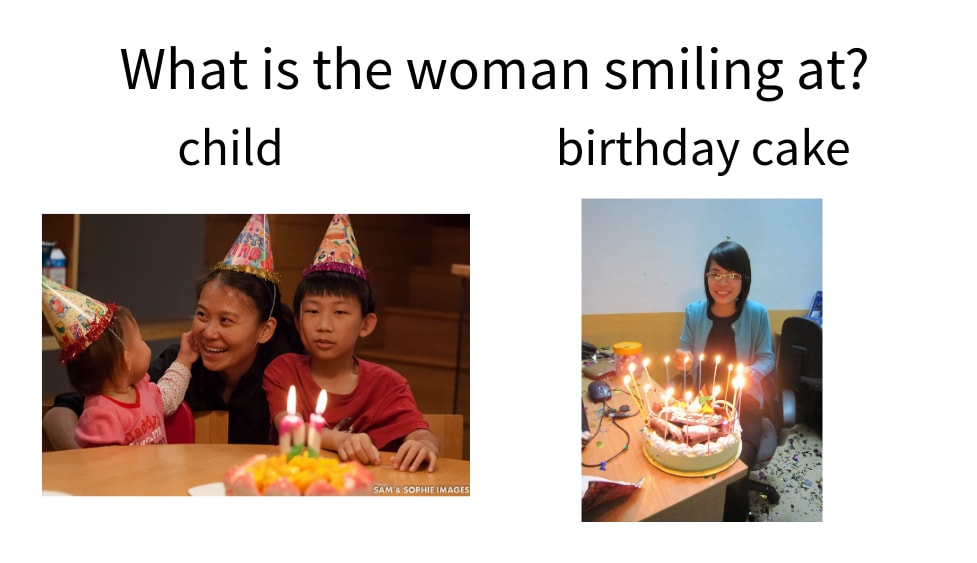} &
\includegraphics[width = 0.22\linewidth]{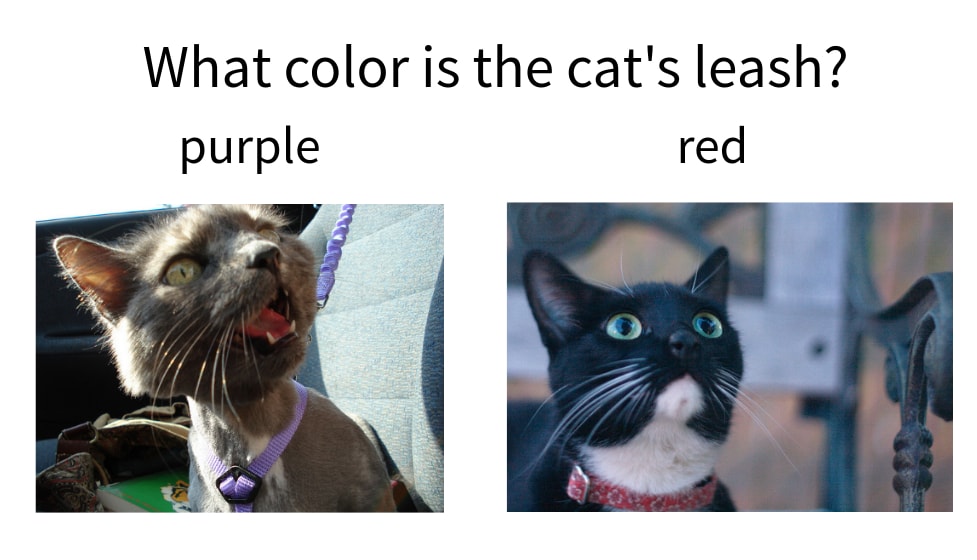} &
\includegraphics[width = 0.22\linewidth]{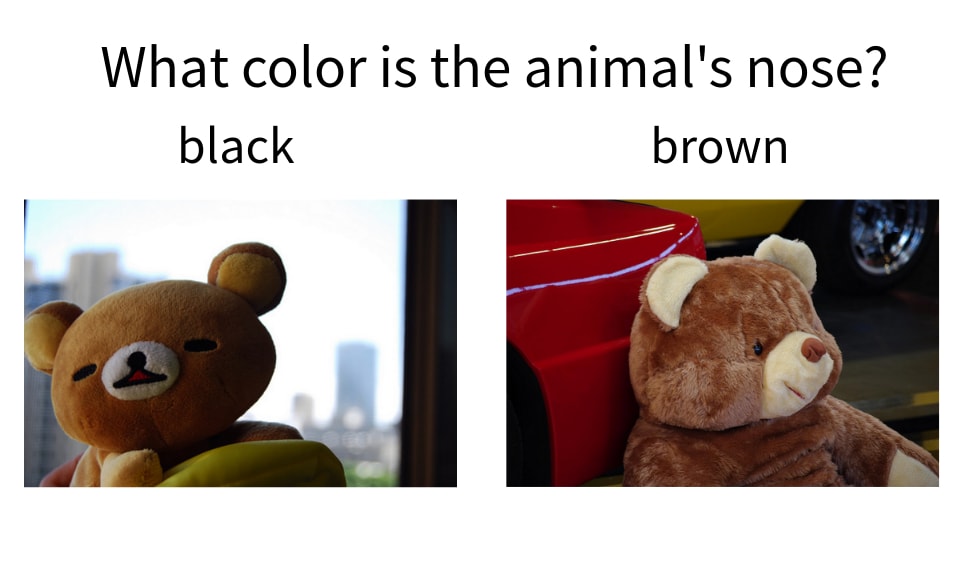}\\
\includegraphics[width = 0.22\linewidth]{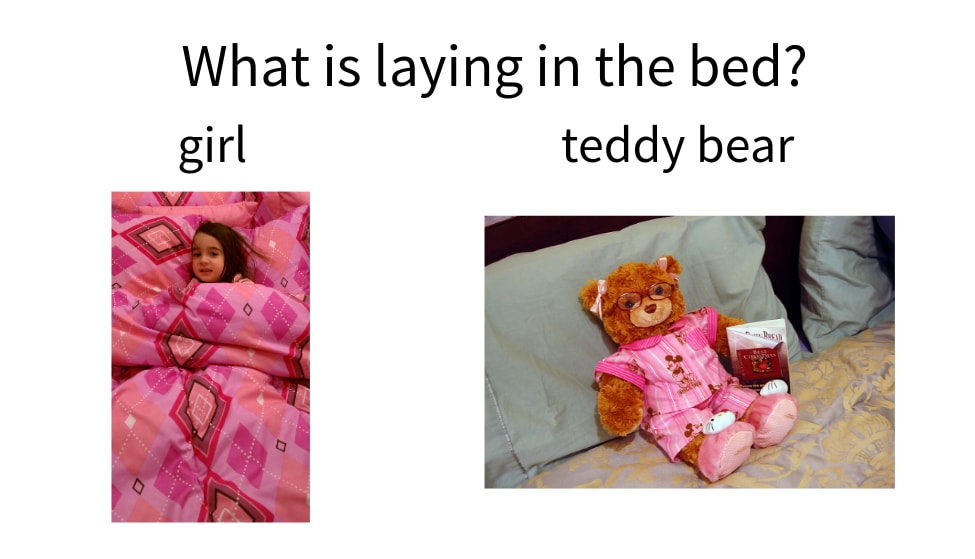} &
\includegraphics[width = 0.22\linewidth]{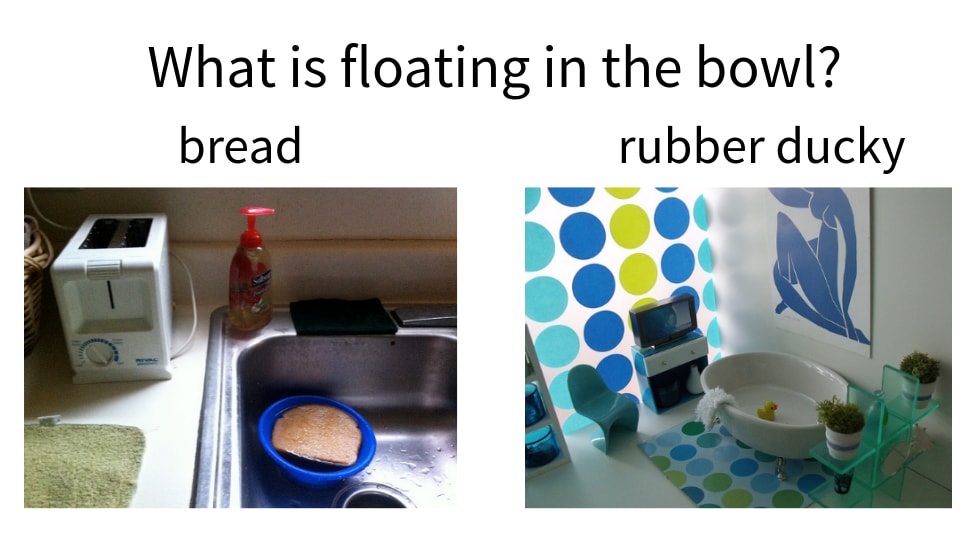} &
\includegraphics[width = 0.22\linewidth]{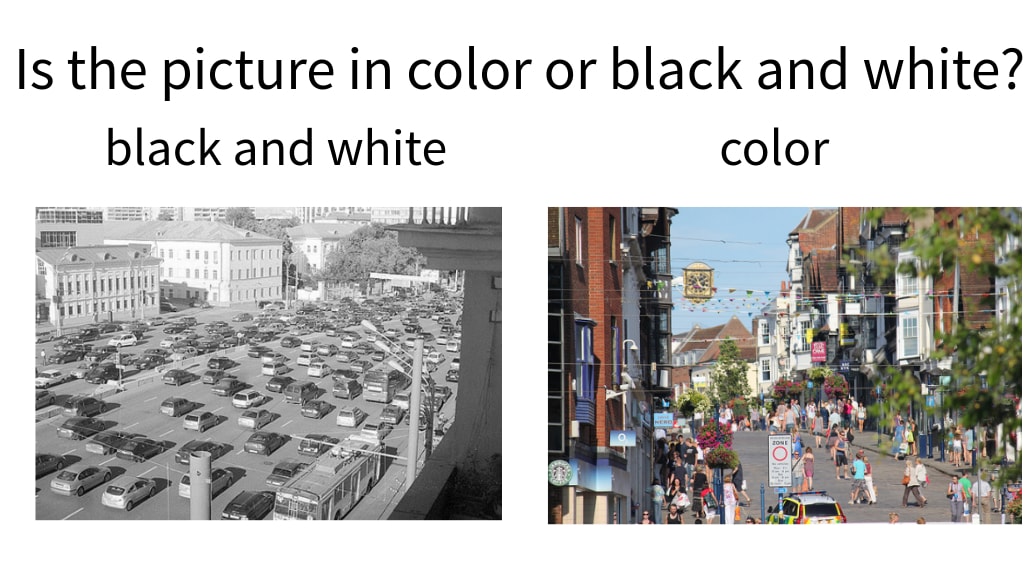} &
\includegraphics[width = 0.22\linewidth]{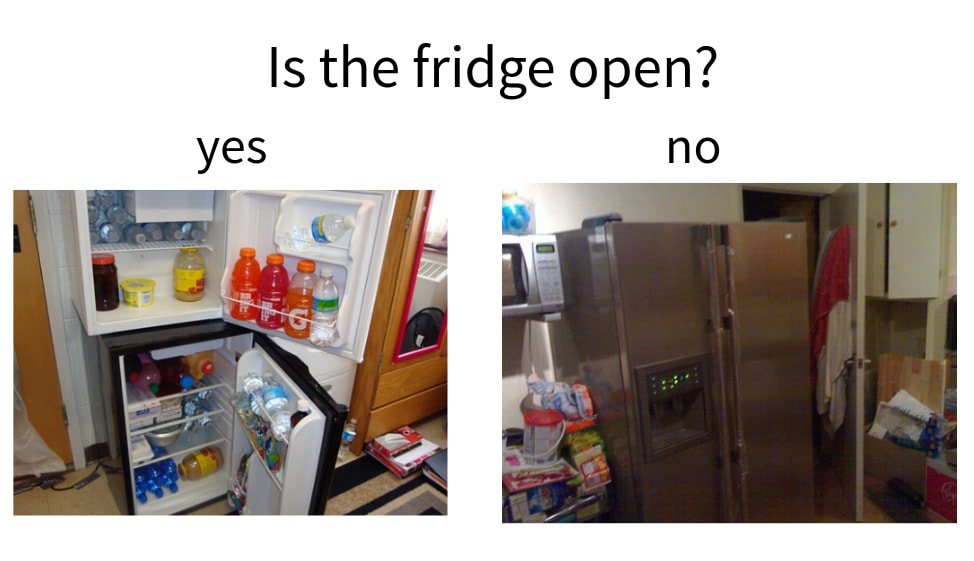}\\
\end{tabular}
\caption{Random examples from our proposed balanced VQA dataset. Each question has two similar images with different answers to the question.}
\label{fig:more_examples}
\end{figure*}

In this work, we propose to counter these language biases and elevate the 
role of image understanding in VQA. In order to accomplish this goal, 
we collect a balanced VQA dataset with significantly reduced language biases. 
Specifically, we create a balanced VQA dataset in the following way -- given 
an (image, question, answer) triplet $(I, Q, A)$ from the VQA dataset, 
we ask a human subject to identify an image $I'$ that is similar to $I$ 
but results in the answer to the question $Q$ to become $A'$ (which is different from $A$). 
Examples from our balanced dataset are shown in \figref{fig:dataset}. 
More random examples can be seen in \figref{fig:more_examples} and on the project website\footnote{\label{website}\url{http://visualqa.org/}}. 

Our hypothesis is that this balanced dataset will force VQA models to focus 
on visual information. 
After all, 
when a question $Q$
has two different answers ($A$ and $A'$) for two different images ($I$ and $I'$ respectively), the only way to know the 
right answer is by looking at the image.
Language-only models have simply no basis for differentiating between the two cases --
$(Q, I)$ and $(Q, I')$, and by construction must get one wrong. 
We believe that this construction will also prevent language+vision models
from achieving high accuracy by exploiting language priors, 
enabling VQA evaluation protocols to more accurately reflect progress in image understanding. 

Our balanced VQA dataset is also particularly difficult because 
the picked complementary image $I'$ is close to the original image $I$ in the 
semantic (fc7) space of VGGNet \cite{Simonyan15} features. Therefore, VQA models 
will need to understand the subtle differences between the two images to predict 
the answers to both the images correctly. 
 
Note that simply ensuring that the answer distribution $P(A)$ is uniform across the 
dataset would not accomplish the goal of alleviating language biases discussed above. 
This is because language models exploit the correlation between question n-grams  
and the answers, \eg questions starting with ``Is there a clock'' has the 
answer ``yes'' 98\% of the time, and questions starting with ``Is the man standing'' 
has the answer ``no'' 69\% of the time. 
What we need is not just higher entropy in $P(A)$ across the dataset, 
but higher entropy in $P(A|Q)$ so that image $I$ must play a role 
in determining $A$. This motivates our balancing on a per-question level. 

Our complete balanced dataset contains approximately \emph{1.1 Million} 
(image, question) pairs -- almost \emph{double} the size of the VQA \cite{VQA} dataset -- 
with approximately \emph{13 Million} associated answers on the $\sim$200k images 
from COCO~\cite{coco}. 
We believe this balanced VQA dataset is a better dataset 
to benchmark VQA approaches, and is publicly available for download on the project website.

Finally, our data collection protocol enables us to develop a counter-example based 
explanation modality. We propose a novel model that not only answers questions 
about images, but also `explains' its answer to an image-question pair by providing 
``hard negatives'' \ie, examples of images that it believes are similar to the image at hand, 
but it believes have different answers to the question. Such an explanation modality will 
allow users of the VQA model to establish greater trust in the model and 
identify its oncoming failures. 

Our main contributions are as follows: 
(1) We balance the existing VQA dataset \cite{VQA} by collecting complementary images 
such that almost every question 
in our balanced dataset is associated with not just a single image, but 
rather a pair of similar images that result in two different answers to the question. 
The result is a more balanced VQA dataset, which is also approximately twice the size 
of the original VQA dataset. 
(2) We evaluate state-of-art VQA models (with publicly available code) on our 
balanced dataset, and show that models trained on the existing `unbalanced' 
VQA dataset perform poorly on our new balanced dataset. This finding confirms 
our hypothesis that these models have been exploiting language priors in the 
existing VQA dataset to achieve higher accuracy. 
(3) 
Finally, our data collection protocol for identifying 
complementary scenes enables us to develop a novel interpretable model, 
which in addition to answering questions about images, also provides a 
counter-example based explanation -- it retrieves images that it believes are 
similar to the original image but have different answers to the question. 
Such explanations can help in building trust for machines among their users.

\section{Related Work}
\label{sec:related_work}

\textbf{Visual Question Answering.} 
A number of recent works have proposed visual question answering datasets 
\cite{VQA, VisualGenome, fritz, Ren_2015_NIPS, baiduVQA, Madlibs, MovieQA,fsvqa} and 
models \cite{MCB, HieCoAtt, NeuralModuleNetworks, dynamic_memory_net_vision, Lu2015, Malinowski_2015_ICCV, binaryVQA,StackedAttentionNetworks,saenko,wang,shih,kim,han,Ilievski,wu,saito,kanan_kafle}.  
Our work builds on top of the VQA dataset from Antol~\etal\cite{VQA}, 
which is one of the most widely used VQA datasets. 
We reduce the language biases present in this popular dataset,  
resulting in a dataset that is more balanced and about twice the size of the VQA dataset. 
We benchmark one `baseline' VQA model \cite{Lu2015}, 
one attention-based VQA model \cite{HieCoAtt}, 
and the winning model from the VQA Real Open Ended Challenge 2016 
\cite{MCB} on our balanced VQA dataset, and compare them to a language-only model. 

\textbf{Data Balancing and Augmentation.}
At a high level, our work may be viewed as constructing a more rigorous evaluation 
protocol by collecting `hard negatives'. In that spirit, it is similar to the 
work of Hodosh~\etal~\cite{julia_cap_bal}, who 
created a binary forced-choice image captioning task, where a machine must 
choose to caption an image with one of two similar captions.  
To compare, Hodosh~\etal~\cite{julia_cap_bal} implemented hand-designed rules 
to create two similar captions for images, 
while we create a novel annotation interface to collect two similar images for questions 
in VQA.

Perhaps the most relevant to our work is that of Zhang~\etal~\cite{binaryVQA}, 
who study this goal of balancing VQA in a fairly restricted setting -- 
binary (yes/no) questions on abstract scenes made from clipart 
(part of the VQA abstract scenes dataset \cite{VQA}). 
Using clipart allows Zhang~\etal to ask human annotators to ``change the clipart scene 
such that the answer to the question changes''. Unfortunately, such fine-grained editing 
of image content is simply not possible in real images. 
The novelty of our work over Zhang~\etal is 
the proposed complementary image data collection interface, application to real images, 
extension to \emph{all} questions (not just binary ones), 
benchmarking of state-of-art VQA models on the balanced dataset, 
and finally the novel VQA model with counter-example based explanations.

\textbf{Models with explanation.}
A number of recent works have proposed mechanisms for generating `explanations' 
\cite{generating_visual_explanations, grad-cam,cam, VQA_interpretability_ICML, sameersingh} 
for the predictions made by deep learning models, which are typically `black-box' 
and non-interpretable. 
\cite{generating_visual_explanations} generates a natural language explanation 
(sentence) for image categories. 
\cite{grad-cam, cam, VQA_interpretability_ICML, sameersingh} provide `visual 
explanations' or spatial maps overlaid on images to highlight the regions that 
the model focused on while making its predictions. 
In this work, we introduce a third explanation modality: counter-examples, 
instances the the model believes are close to but not belonging to the category 
predicted by the model. 

\section{Dataset}
\label{sec:dataset}

We build on top of the VQA dataset introduced by Antol~\etal~\cite{VQA}. 
VQA real images dataset contains just over 204K images from COCO \cite{coco}, 
614K free-form natural language questions (3 questions per image), 
and over 6 million free-form (but concise) answers (10 answers per question). 
While this dataset has spurred significant progress in VQA domain, 
as discussed earlier, it has strong language biases.

\begin{figure}[t]
\centering
\includegraphics[width=1\linewidth]{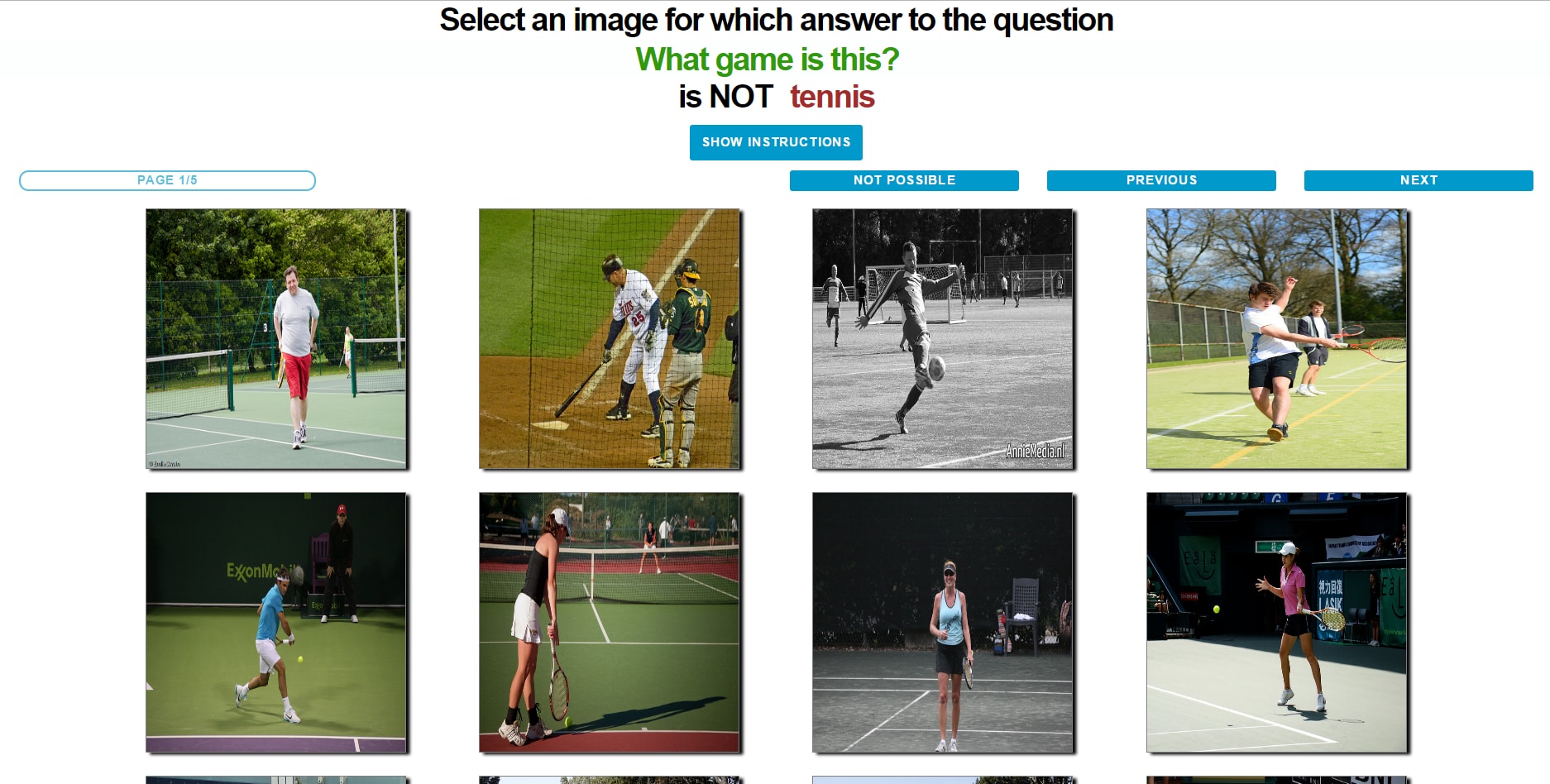}
\caption{A snapshot of our Amazon Mechanical Turk (AMT) interface to collect complementary images.}
\label{fig:interface}
\end{figure}

Our key idea to counter this language bias is the following -- 
for every (image, question, answer) triplet $(I,Q,A)$ in the VQA dataset, 
our goal is to identify an image $I'$ that is similar to $I$, 
but results in the answer to the question $Q$ to become $A'$ 
(which is different from $A$). 
We built an annotation interface (shown in \figref{fig:interface}) 
to collect such complementary images on Amazon Mechanical Turk (AMT). 
AMT workers are shown 24 nearest-neighbor images of $I$, 
the question $Q$, and the answer $A$, and asked to pick an image $I'$ from 
the list of 24 images for which $Q$ ``makes sense'' 
and the answer to $Q$ is \emph{not} $A$. 

To capture ``question makes sense'', 
we explained to the workers (and conducted qualification 
tests to make sure that they understood) that any premise assumed in the 
question must hold true for the image they select. 
For instance, the question ``What is the woman doing?'' assumes that a woman is present 
and can be seen in the image. It does not make sense to ask this question on 
an image without a woman visible in it. 

We compute the 24 nearest neighbors by first representing each image with the 
activations from the penultimate (`fc7') layer of a deep Convolutional 
Neural Network (CNN) -- in particular VGGNet \cite{Simonyan15} -- and then 
using $\ell_2$-distances to compute neighbors. 

After the complementary images are collected, we conduct a second round 
of data annotation to collect answers on these new images. Specifically, 
we show the picked image $I'$ with the question $Q$ to 10 new AMT workers, 
and collect 10 ground truth answers (similar to \cite{VQA}). 
The most common answer among the 10 is the new answer $A'$. 

\begin{figure*}[t]
\centering
\bf{Answers from unbalanced dataset}\par\medskip
\includegraphics[width=\linewidth,height=6cm]{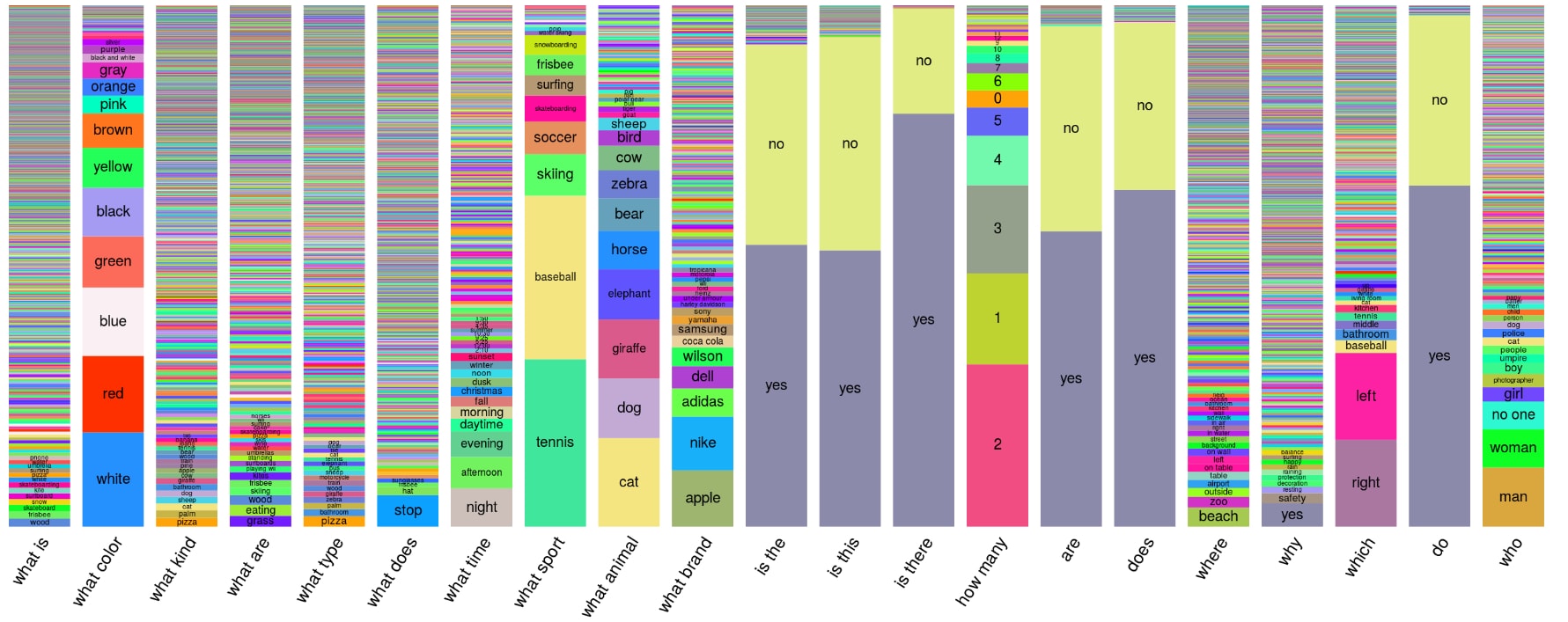}
\bf{Answers from balanced dataset}\par\medskip
\includegraphics[width=\linewidth,height=6cm]{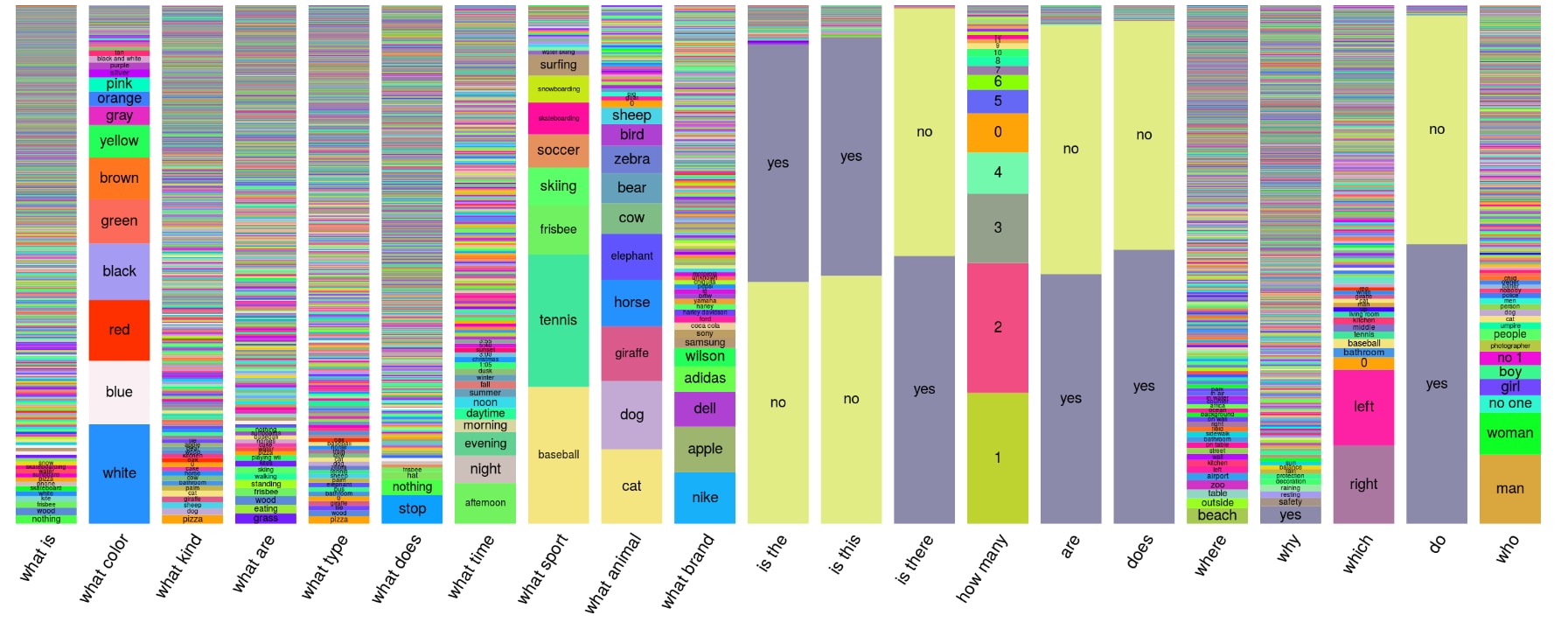}
\caption{Distribution of answers per question type for a random sample of 60K questions from the original (unbalanced) VQA dataset \cite{VQA} (top) and from our proposed balanced dataset (bottom).}
\label{fig:bal_answers}
\end{figure*}

This two-stage data collection process finally results in pairs of complementary 
images $I$ and $I'$ that are semantically similar, but have different answers 
$A$ and $A'$ respectively to the same question $Q$. Since $I$ and $I'$ are 
semantically similar, a VQA model will have to understand the subtle differences 
between $I$ and $I'$ to provide the right answer to both images. 
Example complementary images are shown in \figref{fig:dataset},  \figref{fig:more_examples}, and on the project website.

Note that sometimes it may not be \emph{possible} to pick one of the 24 neighbors as a complementary image. 
This is because either (1) the question does not make sense for any of the 24 images 
(\eg the question is `what is the woman doing?' and none of the neighboring images 
contain a woman), or (2) the question is applicable to some neighboring images, but 
the answer to the question is still $A$ (same as the original image $I$). 
In such cases, our data collection interface allowed AMT 
workers to select ``not possible''. 

We analyzed the data annotated with ``not possible'' selection by AMT workers 
and found that this typically happens when 
(1) the object being talked about in 
the question is too small in the original image and thus the nearest neighbor 
images, while globally similar, do not necessarily contain the object resulting in the question not making sense, or 
(2) when the concept in the question is rare (\eg, when workers are 
asked to pick an image such that the answer to the question 
``What color is the banana?'' is NOT ``yellow''). 

In total, such ``not possible'' selections make up 22\% of all the questions in the VQA dataset.
We believe that a more sophisticated interface that allowed workers to 
scroll through many more than 24 neighboring images could possibly reduce this fraction. 
But, 
(1) it will likely still not be 0 (there may be no image in COCO where the 
answer to ``is the woman flying?'' is NOT ``no''), and 
(2) the task would be significantly more cumbersome for workers, 
making the data collection significantly more expensive. 

We collected complementary images and 
the corresponding new answers for all of train, val and test splits 
of the VQA dataset. 
AMT workers picked ``not possible'' for approximately 135K total 
questions. 
In total, we collected approximately 195K complementary images for train, 93K complementary images for val, and 191K complementary images for test set.
In addition, we augment the test set with $\sim$18K additional (question, image) pairs 
to provide additional means to detect anomalous trends on the test data.
Hence, our complete balanced dataset contains more than 443K train, 214K val and 453K test (question, image) pairs.
Following original VQA dataset \cite{VQA}, we divide our test set into 4 splits: test-dev, test-standard, test-challenge and test-reserve. For more details, please refer to \cite{VQA}.
Our complete balanced dataset is publicly available for download.

We use the publicly released VQA evaluation script in our experiments. 
The evaluation metric uses 10 ground-truth answers for each question to 
compute VQA accuracies. 
As described above, we collected 10 answers for every complementary image 
and its corresponding question to be consistent with the VQA dataset \cite{VQA}.  
Note that while unlikely, it is possible that the majority vote of the 10 new 
answers may not match the intended answer of the person picking the image 
either due to inter-human disagreement, or if the worker selecting the complementary 
image simply made a mistake. 
We find this to be the case -- \ie, $A$ to be the same as $A'$ -- 
for about 9\% of our questions.

\figref{fig:bal_answers} compares the distribution of answers per question-type 
in our new balanced VQA dataset with the original (unbalanced) VQA dataset \cite{VQA}. We notice several interesting trends.
First, binary questions 
(\eg ``is the'', ``is this'', ``is there'', ``are'', ``does'') have a 
\emph{significantly} more balanced distribution over ``yes'' and ``no'' answers 
in our balanced dataset compared to unbalanced VQA dataset.  
``baseball'' is now slightly more popular than ``tennis'' under ``what sport'', 
and more importantly, overall ``baseball'' and ``tennis'' dominate less in the 
answer distribution. Several other sports like ``frisbee'', ``skiing'', 
``soccer'', ``skateboarding'', ``snowboard'' and ``surfing'' are more visible 
in the answer distribution in the balanced dataset, suggesting that it contains 
heavier tails. 
Similar trends can be seen across the board with colors, animals, numbers, \etc. 
Quantitatively, we find that the entropy of answer distributions 
averaged across various question types (weighted by frequency of 
question types) increases by 56\% after balancing, confirming the heavier tails 
in the answer distribution. 

As the statistics show, while our balanced dataset is not perfectly balanced, it is \emph{significantly} more balanced than the original VQA dataset. 
The resultant impact of this balancing on performance of 
state-of-the-art VQA models is discussed in the next section. 

\section{Benchmarking Existing VQA Models}
\label{sec:benchmark}

Our first approach to training a VQA model that emphasizes the 
visual information over language-priors-alone is to re-train 
the existing state-of-art VQA models (with code publicly available 
\cite{Lu2015,HieCoAtt,MCB})
 on our new balanced VQA dataset. 
Our hypothesis is that simply training a model to 
answer questions correctly on our balanced 
dataset will already encourage the model to focus more on the visual signal, 
since the language signal alone has been impoverished. 
We experiment with the following models: 

\textbf{Deeper LSTM Question + norm Image (d-LSTM+n-I)} \cite{Lu2015}: 
This was the VQA model introduced in \cite{VQA} together 
with the dataset. It uses a CNN embedding of the image, 
a Long-Short Term Memory (LSTM) embedding of the question, 
combines these two embeddings via a point-wise multiplication,
followed by a multi-layer perceptron classifier to predict a probability 
distribution over $1000$ most frequent answers in the training dataset. 

\textbf{Hierarchical Co-attention (HieCoAtt)} \cite{HieCoAtt}: 
This is a recent attention-based VQA model that `co-attends' to both the image 
and the question 
to predict an answer. Specifically, it models  
the question (and consequently the image via the co-attention mechanism) 
in a hierarchical fashion: at the word-level, phrase-level and entire question-level. 
These levels are combined recursively to produce a distribution over the  $1000$ most frequent answers. 

\textbf{Multimodal Compact Bilinear Pooling (MCB)} \cite{MCB}: This is 
the winning entry on the real images track of the VQA Challenge 2016. This model 
uses a multimodal compact bilinear pooling mechanism to attend over image features  
and combine the attended image features with language features. These combined 
features are then passed through a fully-connected layer to predict a 
probability distribution over the $3000$ most frequent answers. 
It should be noted that MCB uses image features from a more powerful CNN architecture ResNet \cite{ResNet} while the previous two models use image features from VGGNet \cite{Simonyan15}.

\textbf{Baselines:} To put the accuracies of these models in perspective, 
we compare to the following baselines:  
\textbf{Prior:} Predicting the most common answer in the training set, 
for all test questions. 
The most common answer is ``yes'' in both the unbalanced and balanced sets. 
\textbf{Language-only:} 
This language-only baseline has a similar architecture as Deeper LSTM Question + norm Image \cite{Lu2015} except that it only accepts the question as input and 
does not utilize any visual information. 
Comparing VQA models to language-only ablations quantifies to what 
extent VQA models have succeeded in leveraging the image to answer the 
questions. 

The results are shown in Table~\ref{tab:imbalanced}. 
For fair comparison of accuracies with original (unbalanced) dataset, we create a balanced train set which is of similar size as original dataset (referred to as B$_\text{half}$ in table).  
For benchmarking, we also report results using the full balanced train set.

\begin{table}[h]
\setlength{\tabcolsep}{10pt}
{\small
\begin{center}
\begin{tabular}{@{}lcccc@{}}
\toprule
Approach & UU & UB & B$_\text{half}$B & BB\\
\midrule
Prior &  27.38 & 24.04 & 24.04 & 24.04\\
Language-only & 48.21 &  41.40 &  41.47 & 43.01\\
d-LSTM+n-I \cite{Lu2015}& 54.40  & 47.56 & 49.23 & 51.62\\
HieCoAtt \cite{HieCoAtt} &  57.09 & 50.31 & 51.88 & 54.57\\
MCB \cite{MCB} & 60.36 & 54.22 & 56.08 & 59.14\\
\bottomrule
\end{tabular}
\end{center}
}
\caption {Performance of VQA models when trained/tested on unbalanced/balanced VQA datasets. UB stands for training on \textbf{U}nbalanced train and testing on \textbf{B}alanced val datasets. UU, B$_\text{half}$B and BB are defined analogously.
} 
\label{tab:imbalanced}
\end{table}

We see that the current state-of-art VQA models trained on the original 
(unbalanced) VQA dataset perform significantly worse when evaluated on our 
balanced dataset, compared to evaluating on the original unbalanced VQA dataset 
(\ie, comparing UU to UB in the table). This finding confirms our hypothesis 
that existing models have learned severe language biases present in the dataset, resulting 
in a reduced ability to answer questions correctly when the same question has 
different answers on different images. When these models are trained on our 
balanced dataset, their performance improves (compare UB to B$_\text{half}$B in the table).
Further, when models are trained on complete balanced dataset ($\sim$twice the size of original dataset), the accuracy improves by 2-3\% (compare B$_\text{half}$B to BB). 
This increase in accuracy suggests that current VQA models are data starved, and would benefit from even larger VQA datasets.

As the absolute numbers in the table suggest, there is significant room 
for improvement in building visual understanding models that can extract 
detailed information from images and leverage this information to answer 
free-form natural language questions about images accurately. As expected from the construction of this balanced dataset, 
the question-only approach performs \emph{significantly} worse on the 
balanced dataset compared to the unbalanced dataset, again confirming 
the language-bias in the original VQA dataset, 
and its successful alleviation (though not elimination) in our proposed 
balanced dataset.

Note that in addition to the lack of language bias, visual reasoning is 
also challenging on the balanced dataset since there are pairs of images 
very similar to each other in image representations learned by CNNs, but 
with different answers to the same question. To be successful, VQA models 
need to understand the subtle differences in these images.

The paired construction of our dataset allows us to analyze the performance of 
VQA models in unique ways. Given the prediction of a VQA model, 
we can count the number of questions where \emph{both} 
complementary images ($I$,$I'$) received correct answer predictions for the 
corresponding question $Q$, 
or both received identical (correct or incorrect) answer predictions, 
or both received different answer predictions. 
For the HieCoAtt \cite{HieCoAtt} model, when trained on the unbalanced dataset, 
13.5\% of the pairs were answered correctly, 
59.9\% of the pairs had identical predictions, 
and 40.1\% of the pairs had different predictions. 
In comparison, when trained on balanced dataset, the same model 
answered 17.7\% of the pairs correctly, a 4.2\% increase in performance! 
Moreover, it predicts identical answers for 10.5\% fewer pairs (49.4\%). 
This shows that by training on balanced dataset, this VQA model has 
learned to tell the difference between two otherwise similar images. However, significant room for improvement remains. The VQA model still can not tell the difference between two images that have a noticeable difference -- a difference enough to result in the two images having different ground truth answers for the same question asked by humans.

To benchmark models on VQA v2.0 dataset, we also train these models on VQA v2.0 train+val and report results on VQA v2.0 test-standard in Table~\ref{tab:test_results}.
Papers reporting results on VQA v2.0 dataset are suggested to report test-standard accuracies and compare their methods' accuracies with accuracies reported in Table~\ref{tab:test_results}.

\begin{table}[h]
\setlength{\tabcolsep}{10pt}
{\small
\begin{center}
\begin{tabular}{@{}lcccc@{}}
\toprule
Approach & All & Yes/No & Number & Other\\
\midrule
Prior &  25.98 & 61.20 & 00.36 & 01.17\\
Language-only & 44.26 &  67.01 &  31.55 & 27.37\\
d-LSTM+n-I \cite{Lu2015}& 54.22  & 73.46 & 35.18 & 41.83\\
MCB \cite{MCB} & 62.27 & 78.82 & 38.28 & 53.36\\
\bottomrule
\end{tabular}
\end{center}
}
\caption {Performance of VQA models when trained on VQA v2.0 train+val and tested on VQA v2.0 test-standard dataset.
} 
\label{tab:test_results}
\end{table}


\textbf{Analysis of Accuracies for Different Answer Types:}\\
We further analyze the accuracy breakdown over answer types 
for Multimodal Compact Bilinear Pooling (MCB) \cite{MCB} 
and Hierarchical Co-attention (HieCoAtt) \cite{HieCoAtt} models.

\begin{table}[h]
\setlength{\tabcolsep}{6.8pt}
{\small
\begin{center}
\begin{tabular}{@{}llcccc@{}}
\toprule
Approach & Ans Type & UU & UB & B$_\text{half}$B & BB  \\
\midrule
\multirow{4}{*}{MCB \cite{MCB}} & Yes/No &  81.20  &  70.40  & 74.89 & 77.37\\
&Number & 34.80 & 31.61 &  34.69 & 36.66 \\
&Other& 51.19 & 47.90  & 47.43 & 51.23 \\
&All &  60.36 & 54.22 & 56.08 & 59.14\\
\midrule
\multirow{4}{*}{HieCoAtt \cite{HieCoAtt}} & Yes/No &  79.99 &  67.62  & 70.93 & 71.80\\
&Number & 34.83 &  32.12  & 34.07 & 36.53 \\
&Other& 45.55  & 41.96  & 42.11 & 46.25 \\
&All &  57.09  & 50.31 &  51.88 & 54.57 \\
\bottomrule
\end{tabular}
\end{center}
}
\caption {
Accuracy breakdown over answer types achieved by MCB \cite{MCB} and HieCoAtt \cite{HieCoAtt} models 
when trained/tested on unbalanced/balanced VQA datasets. 
UB stands for training on \textbf{U}nbalanced train and testing on \textbf{B}alanced val datasets. UU, B$_\text{half}$B and BB are defined analogously.
} 
\label{tab:MCB_anstype}
\end{table}

The results are shown in \tableref{tab:MCB_anstype}. First, we immediately notice that the accuracy for the answer-type ``yes/no'' drops significantly from UU to UB  
($\sim$$10.8\%$ for MCB and $\sim$$12.4\%$ for HieCoAtt). This suggests that these VQA 
models are really exploiting language biases for ``yes/no'' type questions, which leads to high 
accuracy on unbalanced val set because the unbalanced val set also contains these biases. 
But performance drops significantly when tested on the balanced val set which has significantly reduced biases.

Second, we note that for both the state-of-art VQA models, the largest source of improvement 
from UB 
to B$_\text{half}$B 
is the ``yes/no'' answer-type ($\sim$$4.5\%$ for MCB and $\sim$$3\%$ for HieCoAtt) and 
the ``number'' answer-type ($\sim$$3\%$ for MCB and $\sim$$2\%$ for HieCoAtt).

This trend is particularly interesting since the ``yes/no'' and ``number'' answer-types are the ones 
where existing approaches have shown minimal improvements. For instance, in the results announced 
at the VQA Real Open Ended Challenge 2016\footnote{\url{http://visualqa.org/challenge.html}}, 
the accuracy gap between the top-4 approaches is a mere $0.15\%$ in ``yes/no'' answer-type category 
(and a gap of $3.48\%$ among the top-10 approaches). 
Similarly, ``number'' answer-type accuracies only vary by $1.51\%$ and $2.64\%$ respectively. 
The primary differences between current generation of state-of-art approaches seem to come from the 
 ``other'' answer-type where accuracies vary by $7.03\%$ and $10.58\%$ among the top-4 and top-10 entries.

This finding suggests that language priors present in the unbalanced VQA dataset 
(particularly in the ``yes/no'' and ``number'' answer-type questions) lead to similar accuracies for all state-of-art VQA models, 
rendering vastly different models virtually indistinguishable from each other (in terms of their accuracies for these answer-types). 
Benchmarking these different VQA models on our balanced dataset (with reduced language priors) 
may finally allow us to distinguish between `good' models (ones that encode the `right' inductive biases for this task, 
such as attention-based or compositional models) from others that are simply high-capacity models tuning themselves to the biases in the dataset.

\section{Counter-example Explanations}
\label{sec:explanation}

We propose a new explanation modality: counter-examples. We propose a model 
that when asked a question about an image, 
not only provides an answer, 
but also provides example images that are similar to the input image but the model 
believes have different answers to the input question. 
This would instill trust in the user that the model does in fact 
`understand' the concept being asked about. 
For instance, for a question ``What color is the fire-hydrant?'' 
a VQA model may be perceived as more trustworthy 
if in addition to saying ``red'', it also adds ``unlike this'' 
and shows an example image containing a fire-hydrant 
that is not red.\footnote{It could easily also convey what color it thinks 
the fire-hydrant is in the counter-example. 
We will explore this in future work.}

\subsection{Model}

Concretely, at test time, our ``negative explanation'' 
or ``counter-example explanation'' model functions in two steps. 
In the first step, similar to a conventional VQA model, 
it takes in an (image, question) pair $(Q, I)$ as input and predicts 
an answer $A_{pred}$. 
In the second step, it uses this predicted answer $A_{pred}$ along with 
the question $Q$ to retrieve an image that is similar to $I$ but has a 
different answer than $A_{pred}$ to the question $Q$. 
To ensure similarity, the model picks one of $K$ nearest neighbor images of $I$,
$I_{NN}=$ \{$I_1$, $I_2$, ..., $I_{K}$\} as the counter-example. 

How may we find these ``negative explanations''? 
One way of picking the counter-example from $I_{NN}$ is to follow the 
classical ``hard negative mining'' strategy popular in computer vision. 
Specifically, simply pick the 
image that has the lowest $P(A_{pred}|Q,I_i)$ where $i \in {1, 2, ..., K}$. 
We compare to this strong baseline. 
While this ensures that $P(A_{pred}|Q,I_i)$ is low for $I_i$, 
it does not ensure that the $Q$ ``makes sense'' for $I_i$. 
Thus, when trying to find a negative explanation for 
``Q: What is the woman doing? A: Playing tennis'', this ``hard negative mining'' 
strategy might pick an image without a woman in it, which would make 
for a confusing and non-meaningful explanation to show to a user, if the goal 
is to convince them that the model has understood the question. 
One could add a component of 
question relevance \cite{ray2016question} to identify better counter-examples.

Instead, we take advantage of our balanced data collection mechanism 
to directly train for identifying a good counter-example. Note that the 
$I'$ picked by humans is a good counter-example, by definition. 
$Q$ is relevant to $I'$ (since workers were asked to ensure it was), 
$I'$ has a different answer $A'$ than $A$ (the original answer),   
and $I'$ is similar to $I$. Thus, we have supervised training data where 
$I'$ is a counter-example from $I_{NN}$ ($K$ = 24) for question $Q$ 
and answer $A$. 
We train a model that learns to provide negative or counter-example 
explanations from this supervised data. 

To summarize, during test time, our model does two things: first it answers 
the question (similar to a conventional VQA model), and second, it explains 
its answer via a counter-example. For the first step, it is given as input 
an image $I$ and a question $Q$, and it outputs a predicted answer $A_{pred}$. 
For the second (explaining) step, it is given as input the question $Q$, 
an answer to be explained $A$\footnote{In practice, this answer to be explained 
would be the answer predicted by the first step $A_{pred}$. However, we only have access to 
negative explanation annotations from humans for the ground-truth answer $A$ to the question. 
Providing $A$ to the explanation module also helps in evaluating the two steps of 
answering and explaining separately.}, 
\emph{and} a set $I_{NN}$ from which the model has to identify the counter-example. 
At training time, the model is given image $I$,
the question $Q$, and the corresponding ground-truth answer $A$ 
to learn to answer questions. 
It is also given $Q$, $A$, $I'$ (human-picked), $I_{NN}$ ($I' \in I_{NN}$) 
to learn to explain.

Our model architecture contains two heads on top of a shared base `trunk'
-- one head for answering the question and the other head for providing 
an explanation. 
Specifically, our model consists of three major components:

\vspace{5pt}

\noindent \textbf{1. Shared base}: 
The first component of our model is learning representations of images and questions. It is a 2-channel 
 network that takes in an image CNN embedding as input in one branch, 
 question LSTM embedding as input in another branch, 
 and combines the two embeddings by a point-wise multiplication. 
 This gives us a joint $QI$ embedding, similar to the model in \cite{Lu2015}.
The second and third components -- the answering model and the explaining model 
-- take in this joint $QI$ embedding as input, 
and therefore can be considered as two heads over this first shared component. 
A total of 25 images -- the original image $I$ and 24 candidate images 
\{$I_1$, $I_2$, ..., $I_{24}$\} are passed through this shared 
component of the network. 

\vspace{5pt}
\noindent \textbf{2. Answering head:} The second component is learning to 
answer questions. Similar to \cite{Lu2015}, it consists of a fully-connected 
layer fed into a softmax that predicts the probability distribution 
over answers given the $QI$ embedding. 
Only the $QI$ embedding corresponding to the original image $I$ 
is passed through this component and result in a cross-entropy loss. 

\vspace{5pt}
\noindent \textbf{3. Explaining head:} The third component is learning to 
explain an answer $A$ 
via a counter-example image. It is a 2-channel network which 
linearly transforms the joint $QI$ embedding (output from the first component) 
and the answer to be explained $A$ (provided as input)\footnote{%
Note that in theory, one \emph{could} provide $A_{pred}$ as input 
during training instead of $A$. After all, this matches the expected use 
case scenario at test time.  
However, this alternate setup (where $A_{pred}$ is provided as input instead of $A$) 
leads to a peculiar and unnatural explanation training goal -- specifically, 
the explanation head will \emph{still be} learning to explain $A$ 
since that is the answer for which we collected negative explanation human annotations. 
It is simply unnatural to build that model that answers a question with 
$A_{pred}$ but learn to explain a different answer $A$! 
Note that this is an interesting scenario where the current push towards 
``end-to-end'' training for everything breaks down. 
} 
into a common 
embedding space. It computes an inner product of these 2 embeddings 
resulting in a scalar number for each image in $I_{NN}$ 
(also provided as input, from which a counter-example is to be picked). 
These $K$ inner-product values for $K$ candidate images are then passed 
through a fully connected layer to generate $K$ scores $S(I_i)$, where 
$i \in \{1, 2, ..., K\}$. The $K$ candidate images \{$I_1$, $I_2$, ..., $I_{K}$\} 
are then sorted according to these scores $S(I_i)$ as being most to least 
likely of being good counter-examples or negative explanations. 
This component is trained with pairwise hinge ranking losses that 
encourage $S(I') - S(I_i) > M - \eps, \quad I_i \in \{I_1, I_2, ..., I_{K}\} \setminus \{I'\}$, 
\ie the score of the human picked image $I'$ is encouraged to be higher 
than all other candidate images by a desired margin of $M$ (a hyperparameter) 
and a slack of $\eps$. This is of course the classical `constraint form' 
of the pairwise hinge ranking loss, and we minimize the standard expression 
$\max\Big(0, M - \big(S(I') - S(I_i)\big)\Big)$. 
The combined loss function for the shared component is
\begin{multline}
\mathcal{L} = -\log P(A|I,Q) \\+ \lambda \sum_{i} 
\max\Big(0, M - \big(S(I') - S(I_i)\big)\Big)
\end{multline}

where, 
the first term is the cross-entropy loss (for training the answering module) 
on $(I,Q)$, 
the second term is the sum of pairwise hinge losses that encourage the 
explaining model to give high score to image $I'$ (picked by humans) 
than other $I_i$s in $I_{NN}$, and  
$\lambda$ is the trade-off weight parameter between the two losses. 

\subsection{Results}
\begin{figure}[t]
\centering
\includegraphics[width=1\linewidth]{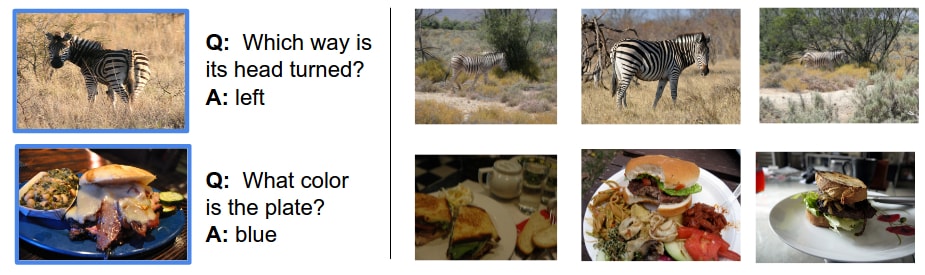}
\caption{Three counter-example or negative explanations (right three columns) generated by our model, along with the input image (left), the input question $Q$ and the predicted answer $A$. 
}
\label{fig:example_negative_explanations}
\end{figure} 

\figref{fig:example_negative_explanations} shows qualitative 
examples of negative explanations produced by our model. We see the original 
image $I$, the question asked $Q$, the answer $A_{pred}$ predicted by 
the VQA head in our model, and top three negative explanations produced 
by the explanation head. We see that most of these explanations are sensible 
and reasonable -- the images are similar to $I$ but with answers that 
are different from those predicted for $I$. 

For quantitative evaluation, we compare our model with a number of baselines:
\textbf{Random:} Sorting the candidate images in $I_{NN}$ randomly. 
That is, a random image from $I_{NN}$ is picked as the most likely 
counter-example. 
\textbf{Distance:} Sorting the candidate images in increasing order 
of their distance from the original image $I$. That is, the image 
from $I_{NN}$ most similar to $I$ is picked as the most likely counter-example. 
\textbf{VQA Model:} Using a VQA model's probability for the predicted 
answer to sort the candidate images in \emph{ascending} order of $P(A|Q,I_i)$. 
That is, the image from $I_{NN}$ \emph{least likely} to have $A$ as the 
answer to $Q$ is picked as the \emph{most likely} counter-example. 

Note that while $I'$ -- the image picked by humans -- is a good counter-example, 
it is not necessarily the unique (or even the ``best'') counter-example. 
Humans were simply asked to pick any image where $Q$ makes sense and the 
answer is not $A$. There was no natural criteria to convey to humans to  
pick the ``best'' one -- it is not clear what ``best'' would mean in the 
first place. To provide robustness to this potential ambiguity in the 
counter-example chosen by humans, in a manner similar to the 
ImageNet \cite{ImageNet} top-5 evaluation metric, we evaluate our approach using the 
Recall@5 metric. It measures how often the human picked $I'$ is among 
the top-5 in the sorted list of $I_i$s in $I_{NN}$ our model produces.

\begin{table}[h]
\setlength{\tabcolsep}{10pt}
{\small
\begin{center}
\begin{tabular}{@{}lcccc@{}}
\toprule
 & Random & Distance & VQA \cite{VQA} & Ours \\
\midrule
 Recall@5  &  20.79 & 42.84 & 21.65 & 43.39  \\
\bottomrule
\end{tabular}
\end{center}
}
\caption {Negative or counter-example explanation performance of our model compared to strong baselines.}
\label{tab:explanation_results}
\end{table}

In \tableref{tab:explanation_results},
we can see that our explanation model significantly outperforms 
the random baseline, as well as the VQA \cite{VQA} model. Interestingly, 
the strongest baseline is Distance. While our approach outperforms it, 
it is clear that identifying an image that is a counter-example to 
$I$ from among $I$'s nearest neighbors is a challenging task. 
Again, this suggests that visual understanding models that can extract 
meaningful details from images still remain elusive.

\section{Conclusion}
\label{sec:conclusion}

To summarize, in this paper we address the strong language priors
for the task of Visual Question Answering and elevate the role of image understanding  
required to be successful on this task. 
We develop a novel data-collection interface to 
`balance' the popular VQA dataset~\cite{VQA} by collecting `complementary' images. For every question in the dataset, we have two complementary images that look similar, but have different answers to the question. 

This effort results in a dataset that is not only more balanced than the original VQA dataset 
by construction, but also is about twice the size. 
We find both qualitatively and quantitatively 
that the `tails' of the answer distribution are heavier in this balanced dataset, 
which reduces the strong language priors that may be exploited by models.
Our complete balanced dataset is publicly available at \url{http://visualqa.org/} as part of the 2nd iteration of the Visual Question Answering Dataset and Challenge (VQA v2.0).

We benchmark a number of (near) state-of-art VQA models 
on our balanced dataset and find that 
testing them on this balanced dataset results in a significant drop in performance, 
confirming our hypothesis that these models 
had indeed exploited language biases. 

Finally, our framework around complementary images 
enables us to develop a novel explainable model -- when asked a question about an image, our 
model not only returns an answer, but also produces a list of similar images that it considers 
`counter-examples', \ie where the answer is not the same as the predicted response. 
Producing such explanations may enable a user to build a better mental model of 
what the system considers a response to mean, and ultimately build trust.  

{
\small
\textbf{Acknowledgements.}
We thank Anitha Kannan and Aishwarya Agrawal for helpful discussions. 
This work was funded in part by NSF CAREER awards to DP
and DB, an ONR YIP award to DP, ONR Grant N00014-14-1-0679 to DB, a Sloan Fellowship to DP, ARO YIP awards
to DB and DP, an Allen Distinguished Investigator award
to DP from the Paul G. Allen Family Foundation, ICTAS
Junior Faculty awards to DB and DP, Google Faculty Research
Awards to DP and DB, Amazon Academic Research
Awards to DP and DB, AWS in Education Research grant
to DB, and NVIDIA GPU donations to DB. The views and
conclusions contained herein are those of the authors and
should not be interpreted as necessarily representing the official policies or endorsements, either expressed or implied,
of the U.S. Government, or any sponsor.
}

{\small
\bibliographystyle{ieee}
\bibliography{references}
}





\end{document}